\documentstyle[twoside,multicol,algorithm,algorithmic,graphicx,lscape,textcomp,url,caption,side,amssymb,amsmath,multirow]{article}

\catcode`\@=11
\long\def\@makefntext#1{
\protect\noindent \hbox to 3.2pt {\hskip-.9pt  
$^{{\footnotesize\@thefnmark}}$\hfil}#1\hfill}		

\def\@makefnmark{\hbox to 0pt{$^{\@thefnmark}$\hss}}	
	
\def\ps@myheadings{%
    \let\@oddfoot\@empty\let\@evenfoot\@empty
    \def\@evenhead{\footnotesize\it\leftmark\hfil}
    \def\@oddhead{\hfil{\footnotesize\it\rightmark}}
    \let\@mkboth\@gobbletwo
    \let\sectionmark\@gobble
    \let\subsectionmark\@gobble
    }

\oddsidemargin=\evensidemargin
\newcounter{sectionc}\newcounter{subsectionc}\newcounter{subsubsectionc}
\renewcommand{\section}[1] {\vspace{14pt}\addtocounter{sectionc}{1}
\setcounter{subsectionc}{0}\setcounter{subsubsectionc}{0}\noindent 
	{\bf\thesectionc. #1}\par\vspace{8pt}}
\renewcommand{\subsection}[1] {\vspace{14pt}\addtocounter{subsectionc}{1}
   \setcounter{subsubsectionc}{0}\noindent 
   {\bf\thesectionc.\thesubsectionc. {\kern1pt \bfit #1}}\par\vspace{8pt}}
\renewcommand{\subsubsection}[1] {\vspace{14pt}
    \addtocounter{subsubsectionc}{1}
	\noindent{\thesectionc.\thesubsectionc.\thesubsubsectionc.
	{\kern1pt \it #1}}\par\vspace{8pt}}
\newcommand{\nonumsection}[1] {\vspace{14pt}\noindent{\bf #1}
	\par\vspace{8pt}}

\newcounter{appendixc}
\newcounter{subappendixc}[appendixc]
\newcounter{subsubappendixc}[subappendixc]
\renewcommand{\thesubappendixc}{\Alph{appendixc}.\arabic{subappendixc}}
\renewcommand{\thesubsubappendixc}
	{\Alph{appendixc}.\arabic{subappendixc}.\arabic{subsubappendixc}}

\renewcommand{\appendix}[1] {\vspace{14pt}
        \refstepcounter{appendixc}
        \setcounter{figure}{0}
        \setcounter{table}{0}
        \setcounter{lemma}{0}
        \setcounter{theorem}{0}
        \setcounter{corollary}{0}
        \setcounter{definition}{0}
        \setcounter{equation}{0}
        \renewcommand{\thefigure}{\Alph{appendixc}.\arabic{figure}}
        \renewcommand{\thetable}{\Alph{appendixc}.\arabic{table}}
        \renewcommand{\theappendixc}{\Alph{appendixc}}
        \renewcommand{\thelemma}{\Alph{appendixc}.\arabic{lemma}}
        \renewcommand{\thetheorem}{\Alph{appendixc}.\arabic{theorem}}
        \renewcommand{\thedefinition}{\Alph{appendixc}.\arabic{definition}}
        \renewcommand{\thecorollary}{\Alph{appendixc}.\arabic{corollary}}
        \renewcommand{\theequation}{\Alph{appendixc}.\arabic{equation}}
        \noindent{\bf Appendix \theappendixc #1}\par\vspace{5pt}}
\newcommand{\subappendix}[1] {\vspace{14pt}
        \refstepcounter{subappendixc}
        \noindent{\bf Appendix \thesubappendixc. {\kern1pt \bfit #1}}
	\par\vspace{8pt}}
\newcommand{\subsubappendix}[1] {\vspace{14pt}
        \refstepcounter{subsubappendixc}
        \noindent{\rm Appendix \thesubsubappendixc. {\kern1pt \it #1}}
	\par\vspace{8pt}}

\newcommand{\textlineskip}{\baselineskip=13pt}
\newcommand{\smalllineskip}{\baselineskip=10pt}

\newcommand{\copyrightheading}[1]
	{\vspace*{-2.5cm}\smalllineskip{\flushleft
	{\footnotesize Preprint submitted to International Journal of Neural Systems, #1}\\
	{\footnotesize The original article is available at \url{https://doi.org/10.1142/S0129065714500087}}\\
	{\footnotesize \copyright\, World Scientific Publishing Company}\\
	 }}

\newcommand{\publisher}[2]{{\begin{center}\tenrm\baselineskip=12pt
	Received #1\\
	Revised #2
	\end{center}
	}}

\def\abstracts#1#2{{
	\centering{\begin{minipage}{5.8in}\small\baselineskip=11pt
	\parindent=0pc #1\par 
	\parindent=2pc #2
	\end{minipage}}\par}}

	{\small\baselineskip=11pt
	 \frenchspacing
	 \begin{list}{\arabic{enumi}.}
        {\usecounter{enumi}\setlength{\parsep}{0pt}    
	 \setlength{\leftmargin 12.7pt}{\rightmargin 0pt}
         \setlength{\itemsep}{0pt} \settowidth
	{\labelwidth}{#1.}\sloppy}}{\end{list}}

\newcounter{itemlistc}
\newcounter{romanlistc}
\newcounter{alphlistc}
\newcounter{arabiclistc}

\newcommand{\fcaption}[1]{
        \refstepcounter{figure}
        \setbox\@tempboxa = \hbox{\small Fig.~\thefigure. #1}
        \ifdim \wd\@tempboxa > 5in
           {\begin{center}
        \parbox{5in}{\small\baselineskip=11pt Fig.~\thefigure. #1}
            \end{center}}
        \else
             {\begin{center}
             {\small Fig.~\thefigure. #1}
              \end{center}}
        \fi}

\newcommand{\tcap}[1]{
        \refstepcounter{table}
        \setbox\@tempboxa = \hbox{\small Table~\thetable. #1}
        \ifdim \wd\@tempboxa > 3.15in
           {\begin{center}
        \parbox{3.15in}{\small\smalllineskip 
	    Table~\thetable. #1}
            \end{center}}
        \else
             {\begin{center}
             {\eightpoint Table~\thetable. #1}
              \end{center}}
        \fi}

\newcommand{\mytcap}[1]{
        \refstepcounter{table}
        \setbox\@tempboxa = \hbox{\small Table~\thetable. #1}
        \ifdim \wd\@tempboxa > 6.30in
           {\begin{center}
        \parbox{6.30n}{\small\smalllineskip 
	    Table~\thetable. #1}
            \end{center}}
        \else
             {\begin{center}
             {\eightpoint Table~\thetable. #1}
              \end{center}}
        \fi}

\def\@citex[#1]#2{\if@filesw\immediate\write\@auxout
	{\string\citation{#2}}\fi
\def\@citea{}\@cite{\@for\@citeb:=#2\do
	{\@citea\def\@citea{,}\@ifundefined
	{b@\@citeb}{{\bf ?}\@warning
	{Citation `\@citeb' on page \thepage \space undefined}}
	{\csname b@\@citeb\endcsname}}}{#1}}

\newif\if@cghi
\def\cite{\@cghitrue\@ifnextchar [{\@tempswatrue
	\@citex}{\@tempswafalse\@citex[]}}
\def\citelow{\@cghifalse\@ifnextchar [{\@tempswatrue
	\@citex}{\@tempswafalse\@citex[]}}
\def\@cite#1#2{{$\null^{#1}$\if@tempswa\typeout
	{IJCGA warning: optional citation argument 
	ignored: `#2'} \fi}}

\def\pmb#1{\setbox0=\hbox{#1}
	\kern-.025em\copy0\kern-\wd0
	\kern.05em\copy0\kern-\wd0
	\kern-.025em\raise.0433em\box0}

\def\fnt#1#2{\footnotetext{\kern-.3em
	{$^{\mbox{\scriptsize #1}}$}{#2}}}

\font\tenrm=cmr10
\font\tenit=cmti10 

\font\bfit=cmbxti10 at 10pt

\font\eightit=cmti8

\def\itlatex{\tenit L\kern-.30em\raise.4ex\hbox{\eightit A}\kern-.14em 
T\kern-.1667em\lower.7ex\hbox{E}\kern-.125em X} 

\def\bsc{{\sc a\kern-7pt\sc a}}
\def\bflatex{\bf L\kern-.30em\raise.3ex\hbox{\bsc}\kern-.18em
T\kern-.1667em\lower.7ex\hbox{E}\kern-.125em X} 

\def\qed{\hbox{${\vcenter{\vbox{			
   \hrule height 0.4pt\hbox{\vrule width 0.4pt height 6pt
   \kern5pt\vrule width 0.4pt}\hrule height 0.4pt}}}$}}

\begin{document}
\setlength{\textheight}{8.78truein}     

\thispagestyle{empty}

\markboth{Multi-Strategy Coevolving Aging Particle Optimization}
{Multi-Strategy Coevolving Aging Particle Optimization}

\textlineskip
\setcounter{page}{1}

\copyrightheading{} 

\vspace*{1.05truein}


\centerline{\large\bf MULTI-STRATEGY COEVOLVING AGING PARTICLE OPTIMIZATION}
\vspace*{0.04truein}
\vspace*{0.45truein}
\centerline{GIOVANNI IACCA}
\vspace*{0.0215truein}
\centerline{\it INCAS\textsuperscript{3}} 
\centerline{\it P.O. Box 797, 9400 AT Assen, The Netherlands}
\centerline{\it E-mail: giovanniiacca@incas3.eu}
\vspace*{14pt}
\centerline{FABIO CARAFFINI and FERRANTE NERI}
\vspace*{0.0215truein}
\centerline{\it Centre for Computational Intelligence (CCI)}
\centerline{\it School of Computer Science and Informatics}
\centerline{\it De Montfort University, The Gateway}
\centerline{\it Leicester LE1 9BH, United Kingdom}
\centerline{\it E-mail: fabio.caraffini@email.dmu.ac.uk, fneri@dmu.ac.uk}

\vspace*{0.3truein}
\publisher{~(to be inserted}{~by Publisher)}

\vspace*{0.29truein}
\abstracts{We propose Multi-Strategy Coevolving Aging Particles (MS-CAP), a novel 
population-based algorithm for black-box optimization. In a memetic fashion, MS-CAP combines 
two components with complementary algorithm logics. In the first stage, each particle is 
perturbed independently along each dimension with a progressively shrinking (decaying) radius, and attracted towards 
the current best solution with an increasing force. In the second phase, the particles are mutated and recombined 
according to a multi-strategy approach in the fashion of the ensemble of mutation strategies in Differential Evolution. The proposed algorithm is 
tested, at different dimensionalities, on two complete black-box optimization benchmarks proposed 
at the Congress on Evolutionary Computation 2010 and 2013. To demonstrate the applicability of the approach, 
we also test MS-CAP to train a Feedforward Neural Network modelling the kinematics of an 8-link robot manipulator. 
The numerical results show that MS-CAP, for the setting considered in this study, tends to outperform the state-of-the-art optimization 
algorithms on a large set of problems, thus resulting in a robust and versatile optimizer.}{}

\vspace*{10pt}\textlineskip

\section{Introduction}\label{sec:intro}
\noindent
The intelligence, from its etymology, is the capability of understanding. When we talk about machine intelligence, the concept takes a slightly different meaning. Since at the moment the humanity has no full knowledge of the brain process of understanding, it cannot be reproduced into a machine. Nonetheless, we can still talk about intelligence of machines if we focus on their ``intelligent behaviour''. More specifically, we can consider the intelligence of a machine as the capability of performing a ``clever choice''. More formally, a choice is clever when it guarantees the best desired conditions. For example, when a route must be decided, the shortest path is in general the choice that guarantees the least fuel consumption (if we neglect the traffic, the pavement of the road etc.). Another example, can be an engineering design, see \cite{bib:Rennera2003} and \cite{bib:Bower2008}. If we consider the design of the profile of some airplane parts, some shape can guarantee the best aerodynamic conditions. A similar consideration can be done for electronic and 
telecommunication problems, see~\cite{bib:Chabuk2012} and~\cite{bib:Tao2012} as well as for management and civil engineering, see ~\cite{bib:Plevris2011} and \cite{bib:Putha2012}.

The problem of performing the ``correct choice'' from an array of (finite or infinite) options is the so-called optimization problem. Optimization is extremely important in engineering and machine learning and has been extensively studied over the last decades. Some pioneering studies on structural optimization have been reported e.g. in~\cite{bib:Adeli1986a},~\cite{bib:Abuyounes1987}, and~\cite{bib:Adeli1988}. The technological progress as well as the needs of the market impose the solution of complex optimization problems that have to deal with many variables, see~\cite{bib:Adeli2000} and \cite{bib:PMS2013}, real-time and hardware limitations, see~\cite{bib:Iacca2013LEGO}, and to tackle both design and control issues, see~\cite{bib:Adeli1998}. Some other approaches integrate human decision within the optimization process, see \cite{bib:Bello-Orgaz2013} and \cite{bib:Rodriguez2010}. Since most of these problems cannot be solved by means of exact methods, meta-heuristics, i.e. algorithms which do not require specific hypotheses on the optimization problem, have been widely diffused. Amongst successful implementations Memetic Computing (MC) approaches, 
i.e. hybrid approaches composed of diverse interacting operators, see~\cite{bib:Moscato1989}, \cite{bib:Neri2012Handbook} and \cite{bib:Neri2012Review}, have been extensively used, see e.g.~\cite{bib:Nguyen2008b},~\cite{bib:Duan2010},~\cite{bib:Ishibuchi2004} and~\cite{bib:Sarma2000}.

It is important to remark that optimization algorithms and neural systems are strictly connected. In order to function properly neural systems need learning. Such learning is de facto the correct selection of a set of parameters characterizing the neural system itself. The choice of these parameters is an optimization problem, see~\cite{bib:Sankari2011},\cite{bib:Amiri2012},\cite{bib:Niceto2012},~\cite{bib:Adeli2009}, and~\cite{bib:Ahmadlou2010}. Since this optimization problem is continuous, usually multivariate, and implicitly noisy, see \cite{bib:Neri2008bcP2P}, algorithms based on Particle Swarm Optimization (PSO) and Differential Evolution (DE) frameworks have been proposed in the literature in several occasions. Some examples of PSO application neural network training is given in \cite{bib:Vilovic2009} and \cite{bib:Li2006a}. As for DE application to neural network training, some successful examples are given in \cite{bib:Ilonen2003} and \cite{bib:Slowik2008}. The problem whether PSO or DE is preferable for neural network training has also been studied in several cases. As expected comparative studies tend to give mixed results and the best choice appears to depend on many factors such as the network size and architecture, see e.g. \cite{bibEspinal2011} and \cite{bib:Garro2011}. 

An important engineering application of the binomial neural systems/optimization algorithms is robotics. A typical application of these methods in this field is the control of robotic arms, with some relevant modern studies presented in~\cite{bib:NeriMcDE2010},~\cite{bib:Niceto2011},~\cite{bib:Orgaz2012},~\cite{bib:Tolu2013}. Another example is modelling: typically, robots are extremely complex devices as for both mechanical and electric aspects, therefore writing a mathematical model of their behaviour can be difficult, see~\cite{bib:Wu2010b}. On the other hand, accurate models can be of great support, if not necessary, for designers and users as they allow a proper understanding and thus control of the robotic devices, see~\cite{bib:Zemalache2007}. These models naturally include multiple parameters that must be identified. The parameter identification is an optimization problem itself as its meaning is the selection of those parameters that guarantee the most reliable fitting (the minimal error) with the 
physical system, see e.g.~\cite{bib:Iacca20123SOME},~\cite{bib:Yiannis2012}, \cite{bib:Nyarko2004}, \cite{bib:Wang2011J}, and~\cite{bib:Liu2012}.

With this background in mind, the present paper proposes a novel, general-purpose MC approach for continuous optimization problems with possible applications in neural systems. The proposed algorithm combines two concurrent/cooperative strategies. The first strategy makes use of the age of each solution to adapt the search logic. The second strategy employs and coordinate multiple perturbation techniques in the fashion of Differential Evolution, see~\cite{bib:Neri2010survey}. The proposed approach is a very versatile implementation as it is able to ensure a very good performance over a wide range of problems including multiple dimensionality values. The comparison with modern algorithms confirms that the proposed scheme is actually able to outperform, on a regular basis, optimization algorithms that represent the state-of-the-art in optimization. In addition, the proposed algorithm has been tested on a real world problem, that is the parameter identification of the kinematic model of a robot manipulator. 
The model in this robotics case is performed with the aid of a Feedforward Neural Network.

The remainder of this paper is organized in the following way. Section 2 describes the principles of the algorithmic functioning and gives details about the implementation. Section 3 shows the numerical result over an extensive testbed of test functions. Section 4 describes the application to robotics of the proposed algorithm. Finally, Section 5 gives the conclusion to this work. 

\section{Multi-Strategy Coevolving Aging Particles}\label{sec:ms-cap}
\noindent
We describe here the proposed Multi-Strategy Coevolving Aging Particles (MS-CAP) algorithm. In the following, we refer without loss of generality to the minimization problem of a fitness function $f(x)$, where $x$ is a candidate solution to the optimization problem at hand, defined in ${\mathbb{R}}^D$, being $D$ the problem size. At the beginning of the algorithm a set (or ``swarm'') of $N$ particles is randomly initialized with a single, randomly sampled initial solution $x_{init}$. Each particle $x_i$ is given a lifetime $life_i$ (initially zero), and, for each j-th variable, a random velocity:
\begin{equation}\label{eq:initVel}
v_{(i,j)} = {\mathcal{U}}(-1/2,1/2) \cdot (ub_j-lb_j) 
\end{equation}
where $ub_j$ and $lb_j$ are respectively the upper and lower bound of the search space along the j-th dimension, and ${\mathcal{U}}(x,y)$ is a number sampled from a uniform distribution in $[x,y)$.

After the initialization phase, the main generational loop of the algorithm starts. Broadly speaking, the MS-CAP can be seen as a Memetic Computing approach composed of two stages: the first one, responsible for coevolving the aging particles, i.e., dispersing and attracting them towards the best solution; the latter, activated when the first one fails at improving upon the best solution, responsible for mutating and recombining the particles according to a multi-strategy mechanism. The two components are detailed in the next two subsections.

\subsection{Coevolving Aging Particles}
\noindent
During the coevolving aging phase, first each particle $x_i$ saves its previous position and fitness; then, it updates, for each j-th variable, its velocity and position according to the following rule:
\begin{equation*}\label{eq:update}
\begin{split}
v_{(i,j)} &= v_{(i,j)} + {\mathcal{U}}(0,1) \cdot \frac{n_{eval}}{max_{eval}} \cdot (x_{(best,j)}-x_{(i,j)})\\
x_{(i,j)} &= x_{(i,j)} + v_{(i,j)}
\end{split}
\end{equation*}
where $n_{eval}$ and $max_{eval}$ represent, respectively, the current and the maximum number of fitness evaluations (the latter is the computational budget allotted to the algorithm), while $x_{best}$ represents the current best solution in the swarm. The meaning of this update rule is that, unlike in classic PSO where the attraction for the global best individual is constant, in the proposed scheme the particles are attracted towards the best solution with a force that progressively increases during the optimization process. Thus, at the beginning the attraction is weak (resulting in a larger exploration pressure), whereas in later stages it becomes stronger. Or, in other words, the update rule becomes more exploitative.

The newly perturbed solution is then evaluated and, in case of improvement upon the best solution, the previous particle is replaced 
and the index of the best particle in the swarm is updated. 

\begin{center}
\begin{scriptsize}
\fbox{\begin{minipage}[ht!]{120mm}
\begin{tabbing}
\quad \= \quad \= \quad \= \quad \= \quad \= \quad \= \kill
     \textbackslash\textbackslash Coevolving Aging Particles\\
     $update = \textbf{false}$\\
     \textbf{for} {$i = 1,2,\dots,N$} \textbf{do}\\
    \> ${x_i}^{old} = x_i$\\
    \> ${f_i}^{old} = f_i$\\
    \> \textbf{for} {$j = 1,2,\dots,D$} \textbf{do}\\
    \>\> $v_{(i,j)} = v_{(i,j)} + {\mathcal{U}}(0,1) \cdot \frac{n_{eval}}{max_{eval}} \cdot (x_{(best,j)}-x_{(i,j)})$\\
    \>\> $x_{(i,j)} = x_{(i,j)} + v_{(i,j)}$\\
    \> \textbf{end for}\\
    \> evaluate $f_i$\\
    \> $n_{eval}=n_{eval}+1$\\
    \> \textbf{if} {$f_i < f_{best}$} \textbf{then}\\
    \>\> $update = \textbf{true}$\\
    \>\> $best = i$\\
    \> \textbf{end if}\\
    \> \textbf{if} {$f_i < {f_i}^{old}$} \textbf{then}\\
    \>\> $life_i = 0$\\
    \> \textbf{else}\\
    \>\> $life_i = life_i+1$\\
    \>\> $decay = e^{-life_i}$\\
    \>\> \textbf{if} {$decay < \varepsilon$} \textbf{then}\\
    \>\>\> $life_i = 0$\\
    \>\>\> $r = {\mathcal{U}}\{\{{1,2,\dots,N}\} - \{i\}\}$\\
    \>\>\> $f_i$ = $f_r$\\
    \>\>\> $v_{(i,j)} = {\mathcal{U}}(-1/2,1/2) \cdot (ub_j-lb_j)$\\
    \>\> \textbf{else}\\
    \>\>\> $x_i = {x_i}^{old}$\\
    \>\>\> $f_i = {f_i}^{old}$\\
    \>\>\> \textbf{if} {$mod(life_i,2) = 0$} \textbf{then}\\
    \>\>\>\> \textbf{for} {$j = 1,2,\dots,D$} \textbf{do}\\
    \>\>\>\>\> $v_{(i,j)} = v_{(i,j)} \cdot (-decay)$\\
    \>\>\>\> \textbf{end for}\\
    \>\>\> \textbf{else}\\
    \>\>\>\> \textbf{for} {$j = 1,2,\dots,D$} \textbf{do}\\
    \>\>\>\>\> $v_{(i,j)} = v_{(i,j)} \cdot (-1)$\\
    \>\>\>\> \textbf{end for}\\
    \>\>\> \textbf{end if}\\
    \>\> \textbf{end if}\\
    \> \textbf{end if}\\
     \textbf{end for}
\end{tabbing}
\end{minipage}
}
\end{scriptsize}
\end{center}
\fcaption{Pseudo-code of Coevolving Aging Particles} \label{alg:CAP}

Additionally, when the perturbed solution improves upon its parent, the particle's lifetime is set to zero. Otherwise, the lifetime is increased by one, and an exponential decay is computed as $decay = e^{-life_i}$. If the decay becomes smaller than a given threshold $\varepsilon$, the particle (as well as its fitness) is replaced with another particle randomly chosen from the swarm, its lifetime is set to zero, and its velocity is reinitialized according to eq.~(\ref{eq:initVel}). If the decay is still larger than the threshold, the perturbed particle (and its fitness) is instead reset to the previous values saved at the beginning of this step. Then, if the condition $mod(life_i,2) = 0$ holds true (being $mod$ the modulo operator), i.e., the lifetime has an even value, the velocity is shrunk in the opposite direction ($v_{(i,j)} = v_{(i,j)} \cdot (-decay)$); otherwise, if the lifetime has an odd value, the magnitude of the velocity is retained, but its sign is changed ($v_{(i,j)} = v_{(i,j)} \cdot (-1)$). As a remark, it should be noted that the maximum age of an ``unsuccessful'' particle, i.e. the number of possible perturbations before being reset, is $\left\lceil-\ln(\varepsilon)\right\rceil$. A pseudo-code of this mechanism is shown in Fig. \ref{alg:CAP}. With $r = {\mathcal{U}}\{\{{1,2,\dots,N}\} - \{i\}\}$, we mean a discrete uniform random number sampled from the set $\{1,2,\dots,N\}$, excluding $\{i\}$.

The rationale behind the Coevolving Aging Particles mechanism is that each particle, while being attracted to the best solution, is also perturbed along each dimension, in alternating directions, and with a progressively shrinking radius which decays exponentially with the ``age'' of the particle. In other words, the particles in the swarm act as micro, parallel local-searchers with an embedded restart mechanism based on the particle decay.

\subsection{Multi-Strategy Mutation and Recombination}
\noindent
Whenever the Coevolving Aging Particles fail at improving upon the current best solution, a further mutation/recombination step is activated in order to exploit the current genetic material and explore the search space with a rich set of moves according to a Differential Evolution logics and inspired by the concept of ensemble in DE schemes, see \cite{bib:ensembleDE}. More specifically, $L$ steps of the following procedure are repeated, where $L$ is a parameter of the algorithm.

At first, each particle $x_i$ saves its previous position and fitness, as in the previous phase. Then, it samples a \emph{scale factor} $F_i$ and a \emph{crossover rate} $CR_i$, respectively from a random uniform distribution in $[0.1,1)$ and $[0,1)$. Finally, the particle selects, randomly and with the same probability, a mutation strategy $mut_i$ and a crossover strategy $xover_i$, respectively from a pool of four mutation and two crossover strategies typical of Differential Evolution. The selected strategy $mut_i$ is then used, with the selected scale factor $F_i$, to generate a mutant solution $x_{mut}$. Using the DE-notation to indicate the mutation strategies\cite{bib:Neri2010survey}, we consider here the following pool $P_{mut}$ of strategies:
\begin{itemize}
   \item rand/1: $x_{mut} = x_r  + F\left( {x_s  - x_t } \right)$
   \item rand/2: $x_{mut} = x_r  + F\left( {x_s  - x_t } \right) + F\left({x_u  - x_v } \right)$
   \item rand-to-best/2: ${x_{mut}} = x_r  + K\left( {x_{best}  - x_i }\right)+ {F\left( {x_r  - x_s } \right)} +{F\left( {x_u  - x_v } \right)}$
   \item cur-to-best/1: ${x_{mut}} = x_i  + F\left( {x_{best}  - x_i }\right) + F\left( {x_s  - x_t } \right)$
\end{itemize}
where the indices $r$, $s$, $t$, $u$ and $v$ are mutually exclusive integers within the range $[1,N]$, randomly generated anew for each i-th mutant solution and also different from the index of the current particle $i$, and $K$ is a parameter randomly chosen in $[0,1]$.

After mutation is applied, the newly mutated particle $x_{mut}$ is recombined with its parent $x_i$ choosing with equal probability a crossover strategy $xover_i$ from a pool of strategies $P_{xover}=\{\text{bin},\text{exp}\}$ consisting of the binomial (or uniform, indicated as ``bin'') and the exponential (or two-point modulo, indicated as ``exp'') strategy\cite{bib:DEbook}, in both cases applied with crossover rate $CR_i$. The recombined solution $x_{xover}$ so generated is then compared with its parent $x_i$ and, in case of improvement, replaces it according to the DE one-to-one spawning logic. 

After $L$ repetitions of this sequence of operations, the particles which were updated (i.e., improved upon) are assigned a new velocity according to eq.~(\ref{eq:initVel}) and their lifetime is set to zero. A pseudo-code of the multi-strategy mutation and recombination component is given in Fig. \ref{alg:MS}.

The algorithm continues applying the Coevolving Aging Particles mechanism described before, until a stop condition based on a maximum budget ($max_{eval}$) is reached. The pseudo-code describing initialization and coordination among the algorithmic components is given in Fig. \ref{alg:mscap2}.

\begin{center}
\begin{scriptsize}
\fbox{\begin{minipage}[ht!]{120mm}
\begin{tabbing}
\quad \= \quad \= \quad \= \quad \= \quad \= \quad \= \kill
     \textbackslash\textbackslash Multi-Strategy Mutation and Recombination\\
    \textbf{for} {$i = 1,2,\dots,N$} \textbf{do}\\
    \> $changed_i = \textbf{false}$\\
    \textbf{end for}\\
    \textbf{for} {$i = 1,2,\dots,L$} \textbf{do}\\
    \> \textbf{for} {$i = 1,2,~\cdots~,N$} \textbf{do}\\
    \>\> ${x_i}^{old} = x_i$\\
    \>\> ${f_i}^{old} = f_i$\\
    \> \textbf{end for}\\
    \> \textbf{for} {$i = 1,2,~\cdots~,N$} \textbf{do}\\
    \>\> $F_i = {\mathcal{U}}(0.1,1)$\\
    \>\> $CR_i = {\mathcal{U}}(0,1)$\\
    \>\> select $mut_i$ in $P_{mut}$\\ 
    \>\> select $xover_i$ in $P_{xover}$\\ 
    \>\> apply $mut_i$ with $F=F_i$ to generate $x_{mut}$\\
    \>\> apply $xover_i$ with $CR=CR_i$ to generate $x_{xover}$\\
    \>\> evaluate $f_{xover}$\\
    \>\> $n_{eval}=n_{eval}+1$\\
    \>\> \textbf{if} {$f_{xover} < {f_i}^{old}$} \textbf{then}\\
    \>\>\> $x_i=x_{xover}$\\
    \>\>\> $f_i=f_{xover}$\\
    \>\>\> $changed_i = \textbf{true}$\\
    \>\> \textbf{end if}\\
    \> \textbf{end for}\\
    \> update $best$\\
    \textbf{end for}\\
    \textbf{for} {$i = 1,2,~\cdots~,N$} \textbf{do}\\
    \> \textbf{if} {$changed_i$} \textbf{then}\\
    \>\> \textbf{for} {$j = 1,2,\dots,D$} \textbf{do}\\
    \>\>\> $v_{(i,j)} = {\mathcal{U}}(-1/2,1/2) \cdot (ub_j-lb_j)$\\
    \>\> \textbf{end for}\\
    \>\> $life_i=0$\\
    \> \textbf{end if}\\
    \textbf{end for}
\end{tabbing}
\end{minipage}
}
\end{scriptsize}
\end{center}
\fcaption{Pseudo-code of the Multi Strategy Component} \label{alg:MS}

\begin{center}
\begin{scriptsize}
\fbox{\begin{minipage}[ht!]{120mm}
\begin{tabbing}
\quad \= \quad \= \quad \= \quad \= \quad \= \quad \= \kill
    \textbackslash\textbackslash initialization\\
    initialize $N$, $\varepsilon$ and $L$\\
    sample an initial solution $x_{init}$ in the search space\\
    $n_{eval}=1$\\
    $best=1$\\
    \textbf{for} {$i = 1,2,\dots,N$} \textbf{do}\\
    \> $x_i = x_{init}$\\
    \> $f_i = f_{init}$\\
    \> $life_i = 0$\\
    \> \textbf{for} {$j = 1,2,\dots,D$} \textbf{do}\\
    \>\> $v_{(i,j)} = {\mathcal{U}}(-1/2,1/2) \cdot (ub_j-lb_j)$\\
    \> \textbf{end for}\\
    \textbf{end for}\\
    $f_{best}=f_1$\\
    \textbackslash\textbackslash main loop\\
    \textbf{while} {\textbf{not} \emph{stop condition}} \textbf{do}\\
    \> \textbackslash\textbackslash Coevolving Aging Particles\\
      \> \textbf{if} \textbf{not} {$update$} \textbf{then}\\
    \> \textbackslash\textbackslash Multi-Strategy Mutation and Recombination\\
    \> \textbf{end if}\\
    \textbf{end while}\\
    \textbf{output} $x_{best}$
\end{tabbing}
\end{minipage}
}
\end{scriptsize}
\end{center}
\fcaption{Pseudo-code of MS-CAP} \label{alg:mscap2}

\subsection{Algorithmic Philosophy}
\noindent
The proposed MS-CAP is a MC approach where two operators perturb a population of candidate solutions from complementary perspectives, see \cite{bib:Krasnogor2004b}. In other words, the two mechanisms perturb the solutions according to two very different logics. The coevolving aging particles perturb the variables separately by means of a randomized mechanism, i.e. performing randomized moves along the axes. On the contrary, the DE-like mutations, use the other points to move diagonally by simultaneously perturbing multiple variables. One of the ideas behind the implementation of MS-CAP is that these two search strategies should complement each other and their alternate use should make the algorithm robust enough to tackle problems with diverse features. More specifically, search operators that perform moves along the axes are suitable for separable problems, see \cite{bib:Ros2008}, while non-separable problems require the use of operators that perform diagonal moves, as e.g. \cite{bib:Hansen2003}. The coevolving population, alternatively perturbed by different operators is supposed to prevent from convergence in proximity of local optima thus handling multimodality in fitness landscapes, see \cite{bib:Prugel-Bennet2010}. Moreover, as highlighted in \cite{bib:Neri2010survey}, DE schemes are characterized by a limited amount of search moves. This effect is compensated in an ensemble fashion by the use of multiple mutation strategies, see \cite{bib:ensembleDE}, and in a memetic fashion by the coevolving particle mechanism. The latter further justifies the support action of the DE scheme as it offers a directional search led by the best particle and appears to have a crucially beneficial effect on the DE scheme, as shown in \cite{bib:Caponio2009a}. The aging mechanism is also very important within this scheme as it allows a natural refreshment of the available search directions. More specifically, both PSO and DE like schemes can be prone to stagnation, see \cite{bib:Weber2010SOCO}, and a refresh action appears to offer a (partial) restart to the search, thus allowing the optimization process to successfully progress, see also \cite{bib:NeriDEcDE2011}. A refreshment mechanism is typical in the context of many optimization algorithms. Also aging mechanisms have been recently proposed in some optimization methods: for instance, 
in \cite{bib:Chen2013} aging is applied within a PSO framework to update the global best. However, while in that mechanism the aging of the global best is used to allow that the other particles improve upon its performance, in the present paper the aging mechanism aims at refreshing the entire population while it is perturbed by two concurrent sets of perturbation rules. All in all, MS-CAP has been designed to be, within the respect of the No Free Lunch Theorem \cite{bib:Wolpert1997}, a robust scheme whose strength is within a proper balancing of diverse components that have the role of compensating each other and aim at achieving solutions with a high quality. 

\section{Evaluation on Benchmark Problems}\label{sec:benchmark}
\noindent
In order to assess the performance of MS-CAP on a broad set of real-parameter optimization problems, we evaluate the results obtained by the proposed algorithm on two different benchmarks, namely:
\begin{itemize}
 \item the benchmark used at the CEC 2013\cite{bib:cec2013}, composed of $28$ bound-constrained test functions;
 \item the large-scale optimization benchmark used at CEC 2010\cite{bib:CEC2010}, composed of $20$ bound-constrained test functions.
\end{itemize}
Furthermore, we study the scalability properties of the proposed algorithm testing the CEC 2013 benchmark in $10$, $30$ and $50$ dimensions, and the CEC 2010 benchmark in $1000$ dimensions. 

To have a heterogeneous comparison, we confront MS-CAP with ten state-of-the-art optimization algorithms which make use of different search logics and algorithmic structures, namely (1) algorithms whose structure is based on Differential Evolution, and (2) what we call here ``alternative'' meta-heuristics, i.e. methods based on PSO, Evolution Strategy, and memetic computing. The comparative setup can be summarized as follows:
\\\\
\emph{Differential Evolution based algorithms}\\
  \begin{itemize}
    \item Self Adaptive Differential Evolution (SADE)\cite{bib:Qin2009}, with Learning Period $LP=20$ and population size $N_p=50$;
    \item Adaptive Differential Evolution (JADE)\cite{bib:Zhang2009}, with population size equal to $60$ individuals, $p=0.05$ and adaptation rate $c=0.1$;
    \item Self Adaptive Parameters in Differential Evolution (jDE)\cite{bib:Brest2006a}, with $F_l=0.1$, $F_u=0.9$, $\tau_1=\tau_2=0.1$ and population size $N_p=50$;
    \item Modified Differential Evolution + pBX crossover (MDE-pBX)\cite{bib:das&sugantah2012}, with population size equal to $100$ individuals and group size $q$ equal to $15\%$ of the population size.
    \item Ensemble of Parameters and Strategies in Differential Evolution (EPSDE)\cite{bib:ensembleDE,bib:suganthan2010}, with $N_p=50$, parameter pools $P_{CR}=\{0.1, 0.5, 0.9\}$ and $P_F=\{0.5, 0.9\}$, and pools of strategies $P_{xover}=\{\text{bin},\text{exp}\}$ and $P_{mut}=\{\text{cur-to-pbest/1},\text{cur-to-rand/1}\}$.
  \end{itemize}
\vspace{1em}
\emph{Alternative meta-heuristics}\\
  \begin{itemize}
   \item Comprehensive Learning Particle Swarm Optimizer (CLPSO)\cite{bib:Liang2006}, with population size equal to $60$ individuals;
   \item Cooperatively Coevolving Particle Swarms Optimizer (CCPSO2)\cite{bib:xiaodongli2012}, with population size equal to $30$ individuals, Cauchy/Gaussian sampling selection probability $p=0.5$ and set of potential group sizes $S=\{2, 5\}$, $S=\{2, 5, 10\}$, $S=\{2, 5, 10, 25\}$, for experiments in $10$, $30$ and $50$ dimensions, respectively;
   \item Parallel Memetic Structures (PMS)\cite{bib:PMS2013}, with $\alpha_e=0.95$, $\rho=0.4$, $150$ iterations for short distance exploration, and Rosenbrock tolerance $\varepsilon=10^{-5}$;
   \item Memetic Algorithm with CMA-ES Chains (MA-CMA-Chains) proposed in\cite{bib:Molina2010b} with population size equal to $60$ individuals, probability of updating a chromosome by mutation equal to $0.125$, local/global search ratio $r_{L/G}=0.5$, BLX-$\alpha$ crossover with $\alpha=0.5$, $n_{ass}$ parameter for Negative Assortative Mating set to $3$, LS intensity stretch $I_{str}=500$ and threshold $\delta^{min}_{LS}=10^{-8}$;
   \item Covariance Matrix Adaptation Evolution Strategy (CMA-ES)\cite{bib:Hansen2003}, with the default parameter setting of the original implementation\cite{web:cmaes}, namely $\lambda=\lfloor4 + 3\ln(D)\rfloor$, $\mu=\lfloor\lambda/2\rfloor$, and initial step-size $\sigma=0.2$.
  \end{itemize}
In $1000$ dimensions, the experiments have been carried out by replacing MA-CMA-Chains with its corresponding large scale variant\cite{bib:LozanoSI2011}:
\begin{itemize}
\item Memetic Algorithm with Subgrouping Solis Wets Chains (MA-SSW-Chains) proposed in\cite{bib:Molina2010a} with population size equal to $100$ individuals, probability of updating a chromosome by mutation equal to $0.125$, local/global search ratio $r_{L/G}=0.5$, BLX-$\alpha$ crossover with $\alpha=0.5$, $n_{ass}$ parameter for Negative Assortative Mating set to $3$, LS intensity stretch $I_{str}=500$ and threshold $\delta^{min}_{LS}=0$.
\end{itemize}
In the following, we indicate with ``MACh'' either MA-CMA-Chains or MA-SSW-Chains, depending on the problem dimension. 

For each algorithm and test function, we execute $100$ independent runs, each one with a computational budget of $5000~\times~D$ fitness evaluations (where $D$ is the problem dimension). To handle the search space bounds, we implement in all the algorithms a toroidal mechanism, consisting of the following: given an interval $\left[a,b\right]$, if $x_i=b+\zeta$, i.e. the i-th design variable exceeds the upper bound by a quantity $\zeta$, its value is replaced with $a+\zeta$. A similar mechanism is applied for the lower bound. The entire experimental setup (fitness functions and algorithms) is coded in Java and executed on a hybrid network composed of Linux and Mac computers, using the distributed optimization platform Kimeme\cite{web:kimeme}. 

As a final remark, it should be noted that each algorithm is executed with the parameter setting suggested in its seminal paper. As for MS-CAP, we set $N_p=50$, $\varepsilon=10^{-6}$ (corresponding to a maximum lifetime of $14$) and $L=3$. The analysis of parameter sensitivity, discussed in subsection 3.2, revealed that this setting guarantees the best trade-off in terms of optimization and scalability at different dimensions. 

Tables~\ref{tab:res10a}-\ref{tab:res1000b} show, for each test function and problem dimension, the mean and the standard deviation (over $100$ runs) of the fitness error (with respect to the global optimum) obtained by MS-CAP, jDE, CMA-ES, and CCPSO2 at the end of the allotted budget. For the sake of brevity, we report the numerical results of only four algorithms, chosen as representative set. The entire set of detailed results is available at the link \url{https://sites.google.com/site/facaraff/home/Downloads/MS-CAP_Detailed_Results.pdf}.

In the same tables, we report next to each fitness error the result of each pair-wise statistical comparison between the fitness errors obtained by MS-CAP (taken as reference) and those obtained with the algorithm in the corresponding column name. In symbols, ``='' indicates an equivalent performance, while ``+'' (``-'') indicates that MS-CAP has a better (worse) performance, with respect to the algorithm in the column label, i.e., it shows a smaller (larger) fitness error. 

The statistical comparison is conducted as follows: first, we verify the normality of the two distributions with the Shapiro-Wilk test\cite{bib:ShapiroWilk}; if both samples are normally distributed, we then test the homogeneity of their variances (homoscedasticity) with an F-test\cite{bib:SnedCoch}. If variances are equal, we compare the two distributions by means of the Student t-test\cite{bib:student08ttest}, otherwise we adopt the t-test variant proposed by Welch\cite{bib:Welch}. More specifically, we first test the null-hypothesis of equal distributions (i.e., the two algorithms under comparison are statistically equivalent from an optimization point of view); then, we test the null-hypothesis that the fitness errors of the reference algorithm (MS-CAP) are statistically smaller than those obtained with the algorithm under comparison. In case of non-normal distributions, we instead test the null-hypotheses by means of the non-parametric Wilcoxon Rank-Sum test\cite{bib:Wilcoxon1945}. In all the tests, we consider a confidence level of $0.95$ ($\alpha=0.05$).
\begin{table*}[!ht]
\begin{center}
\captionsetup{justification=centering,width=22cm,margin=0pt,margin=0pt}
\caption{\label{tab:res10a} Statistical comparison of MS-CAP against jDE, CMA-ES, and CCPSO2 on CEC 2013 in $10$ dimensions}
\vspace{-0.3cm}
\begin{tiny}
\begin{tabular}{l|r@{$\,\pm\,$}l|c|r@{$\,\pm\,$}l|c|r@{$\,\pm\,$}l|c|r@{$\,\pm\,$}l|c}
\hline\hline
              &      \multicolumn{3}{c|}{MS-CAP}    &   \multicolumn{3}{c|}{jDE}    &   \multicolumn{3}{c|}{CMA-ES}&   \multicolumn{3}{c}{CCPSO2}\\
\hline
$f_{1}$ & \multicolumn{3}{c|}{$\mathbf{0.00e+00}\,\pm\,\mathbf{0.00e+00}$} & $0.00e+00$ & $0.00e+00$ & + & $0.00e+00$ & $0.00e+00$ & + & $3.08e-03$ & $1.05e-02$ & +\\
$f_{2}$ & \multicolumn{3}{c|}{$2.40e+03\,\pm\,6.42e+03$} & $7.96e+03$ & $1.07e+04$ & + & $\mathbf{0.00e+00}$ & $\mathbf{0.00e+00}$ & - & $1.80e+06$ & $1.21e+06$ & +\\
$f_{3}$ & \multicolumn{3}{c|}{$1.29e+03\,\pm\,6.30e+03$} & $1.07e+02$ & $7.02e+02$ & = & $\mathbf{7.69e-02}$ & $\mathbf{6.40e-01}$ & - & $7.41e+07$ & $1.12e+08$ & +\\
$f_{4}$ & \multicolumn{3}{c|}{$4.56e+00\,\pm\,7.11e+00$} & $4.75e+01$ & $6.54e+01$ & + & $\mathbf{0.00e+00}$ & $\mathbf{0.00e+00}$ & - & $1.05e+04$ & $2.69e+03$ & +\\
$f_{5}$ & \multicolumn{3}{c|}{$\mathbf{0.00e+00}\,\pm\,\mathbf{0.00e+00}$} & $0.00e+00$ & $0.00e+00$ & + & $0.00e+00$ & $0.00e+00$ & + & $2.20e-02$ & $6.13e-02$ & +\\
$f_{6}$ & \multicolumn{3}{c|}{$5.00e+00\,\pm\,4.91e+00$} & $5.60e+00$ & $4.85e+00$ & + & $6.95e+00$ & $8.44e+00$ & - & $\mathbf{4.67e+00}$ & $\mathbf{7.85e+00}$ & +\\
$f_{7}$ & \multicolumn{3}{c|}{$1.37e+00\,\pm\,6.15e+00$} & $\mathbf{9.87e-02}$ & $\mathbf{1.21e-01}$ & - & $6.36e+13$ & $6.32e+14$ & + & $3.99e+01$ & $1.26e+01$ & +\\
$f_{8}$ & \multicolumn{3}{c|}{$\mathbf{2.03e+01}\,\pm\,\mathbf{1.25e-01}$} & $2.04e+01$ & $7.17e-02$ & + & $2.04e+01$ & $1.16e-01$ & + & $2.04e+01$ & $7.48e-02$ & +\\
$f_{9}$ & \multicolumn{3}{c|}{$\mathbf{2.64e+00}\,\pm\,\mathbf{1.36e+00}$} & $4.34e+00$ & $2.16e+00$ & + & $1.51e+01$ & $4.02e+00$ & + & $5.48e+00$ & $8.99e-01$ & +\\
$f_{10}$ & \multicolumn{3}{c|}{$8.70e-02\,\pm\,5.89e-02$} & $2.64e-01$ & $8.70e-02$ & + & $\mathbf{1.60e-02}$ & $\mathbf{1.36e-02}$ & - & $1.93e+00$ & $9.27e-01$ & +\\
$f_{11}$ & \multicolumn{3}{c|}{$0.00e+00\,\pm\,5.68e-15$} & $\mathbf{0.00e+00}$ & $\mathbf{0.00e+00}$ & = & $2.56e+02$ & $2.89e+02$ & + & $2.76e+00$ & $1.85e+00$ & +\\
$f_{12}$ & \multicolumn{3}{c|}{$\mathbf{1.04e+01}\,\pm\,\mathbf{4.00e+00}$} & $1.95e+01$ & $4.03e+00$ & + & $3.30e+02$ & $3.15e+02$ & + & $3.39e+01$ & $1.02e+01$ & +\\
$f_{13}$ & \multicolumn{3}{c|}{$\mathbf{1.42e+01}\,\pm\,\mathbf{7.30e+00}$} & $1.85e+01$ & $4.82e+00$ & + & $2.29e+02$ & $2.76e+02$ & + & $4.22e+01$ & $8.88e+00$ & +\\
$f_{14}$ & \multicolumn{3}{c|}{$1.77e+01\,\pm\,1.27e+02$} & $\mathbf{1.40e+01}$ & $\mathbf{9.46e+00}$ & + & $1.78e+03$ & $4.21e+02$ & + & $8.67e+01$ & $6.15e+01$ & +\\
$f_{15}$ & \multicolumn{3}{c|}{$\mathbf{7.56e+02}\,\pm\,\mathbf{2.40e+02}$} & $1.31e+03$ & $1.72e+02$ & + & $1.78e+03$ & $4.00e+02$ & + & $1.03e+03$ & $2.70e+02$ & +\\
$f_{16}$ & \multicolumn{3}{c|}{$\mathbf{2.62e-01}\,\pm\,\mathbf{1.72e-01}$} & $1.23e+00$ & $2.32e-01$ & + & $3.90e-01$ & $3.24e-01$ & + & $1.31e+00$ & $2.35e-01$ & +\\
$f_{17}$ & \multicolumn{3}{c|}{$\mathbf{1.04e+01}\,\pm\,\mathbf{1.42e-01}$} & $1.16e+01$ & $4.52e-01$ & + & $9.74e+02$ & $3.03e+02$ & + & $1.79e+01$ & $2.64e+00$ & +\\
$f_{18}$ & \multicolumn{3}{c|}{$\mathbf{2.08e+01}\,\pm\,\mathbf{5.43e+00}$} & $3.74e+01$ & $4.11e+00$ & + & $1.03e+03$ & $3.15e+02$ & + & $5.82e+01$ & $6.30e+00$ & +\\
$f_{19}$ & \multicolumn{3}{c|}{$\mathbf{3.99e-01}\,\pm\,\mathbf{1.32e-01}$} & $8.60e-01$ & $1.39e-01$ & + & $1.18e+00$ & $4.76e-01$ & + & $1.00e+00$ & $3.69e-01$ & +\\
$f_{20}$ & \multicolumn{3}{c|}{$\mathbf{3.02e+00}\,\pm\,\mathbf{5.87e-01}$} & $3.05e+00$ & $2.96e-01$ & = & $4.79e+00$ & $2.72e-01$ & + & $3.59e+00$ & $2.16e-01$ & +\\
$f_{21}$ & \multicolumn{3}{c|}{$3.78e+02\,\pm\,7.30e+01$} & $3.96e+02$ & $2.80e+01$ & + & $3.87e+02$ & $5.04e+01$ & = & $\mathbf{3.68e+02}$ & $\mathbf{6.68e+01}$ & +\\
$f_{22}$ & \multicolumn{3}{c|}{$\mathbf{7.08e+01}\,\pm\,\mathbf{1.07e+02}$} & $2.28e+02$ & $7.57e+01$ & + & $2.32e+03$ & $4.07e+02$ & + & $1.23e+02$ & $6.60e+01$ & +\\
$f_{23}$ & \multicolumn{3}{c|}{$\mathbf{9.79e+02}\,\pm\,\mathbf{2.99e+02}$} & $1.50e+03$ & $1.94e+02$ & + & $2.24e+03$ & $4.28e+02$ & + & $1.37e+03$ & $2.82e+02$ & +\\
$f_{24}$ & \multicolumn{3}{c|}{$1.95e+02\,\pm\,2.47e+01$} & $\mathbf{1.88e+02}$ & $\mathbf{2.80e+01}$ & - & $3.73e+02$ & $1.36e+02$ & + & $2.11e+02$ & $1.80e+01$ & +\\
$f_{25}$ & \multicolumn{3}{c|}{$\mathbf{1.98e+02}\,\pm\,\mathbf{1.42e+01}$} & $1.98e+02$ & $1.29e+01$ & - & $2.61e+02$ & $5.29e+01$ & + & $2.12e+02$ & $1.46e+01$ & +\\
$f_{26}$ & \multicolumn{3}{c|}{$1.31e+02\,\pm\,3.32e+01$} & $\mathbf{1.29e+02}$ & $\mathbf{1.93e+01}$ & + & $2.57e+02$ & $1.09e+02$ & + & $1.71e+02$ & $2.37e+01$ & +\\
$f_{27}$ & \multicolumn{3}{c|}{$3.23e+02\,\pm\,4.47e+01$} & $\mathbf{3.06e+02}$ & $\mathbf{2.23e+01}$ & - & $4.01e+02$ & $9.94e+01$ & + & $4.33e+02$ & $5.71e+01$ & +\\
$f_{28}$ & \multicolumn{3}{c|}{$\mathbf{2.86e+02}\,\pm\,\mathbf{5.15e+01}$} & $2.86e+02$ & $5.10e+01$ & - & $1.22e+03$ & $1.13e+03$ & + & $4.01e+02$ & $1.63e+02$ & +\\
\hline\hline
\end{tabular}
\end{tiny}
\end{center}
\end{table*}
\begin{table*}[!ht]
\begin{center}
\captionsetup{justification=centering,width=22cm,margin=0pt,margin=0pt}
\caption{\label{tab:res30a} Statistical comparison of MS-CAP against jDE, CMA-ES, and CCPSO2 on CEC 2013 in $30$ dimensions}
\vspace{-0.3cm}
\begin{tiny}
\begin{tabular}{l|r@{$\,\pm\,$}l|c|r@{$\,\pm\,$}l|c|r@{$\,\pm\,$}l|c|r@{$\,\pm\,$}l|c}
\hline\hline
              &      \multicolumn{3}{c|}{MS-CAP}    &   \multicolumn{3}{c|}{jDE}    &   \multicolumn{3}{c|}{CMA-ES}&   \multicolumn{3}{c}{CCPSO2}\\
\hline
$f_{1}$ & \multicolumn{3}{c|}{$0.00e+00\,\pm\,1.41e-13$} & $0.00e+00$ & $7.19e-14$ & - & $\mathbf{0.00e+00}$ & $\mathbf{1.18e-13}$ & = & $1.36e-12$ & $6.01e-12$ & +\\
$f_{2}$ & \multicolumn{3}{c|}{$4.02e+05\,\pm\,2.10e+05$} & $2.78e+06$ & $1.46e+06$ & + & $\mathbf{0.00e+00}$ & $\mathbf{1.54e-13}$ & - & $2.14e+06$ & $1.04e+06$ & +\\
$f_{3}$ & \multicolumn{3}{c|}{$5.06e+06\,\pm\,6.60e+06$} & $1.88e+06$ & $2.78e+06$ & - & $\mathbf{9.24e+01}$ & $\mathbf{4.00e+02}$ & - & $1.13e+09$ & $1.18e+09$ & +\\
$f_{4}$ & \multicolumn{3}{c|}{$1.72e+03\,\pm\,9.10e+02$} & $7.76e+03$ & $2.59e+03$ & + & $\mathbf{0.00e+00}$ & $\mathbf{1.29e-13}$ & - & $5.64e+04$ & $2.09e+04$ & +\\
$f_{5}$ & \multicolumn{3}{c|}{$\mathbf{1.14e-13}\,\pm\,\mathbf{1.61e-14}$} & $1.14e-13$ & $1.61e-14$ & - & $9.09e-13$ & $2.46e-12$ & + & $3.04e-07$ & $8.74e-07$ & +\\
$f_{6}$ & \multicolumn{3}{c|}{$2.41e+01\,\pm\,2.18e+01$} & $1.94e+01$ & $1.59e+01$ & - & $\mathbf{4.83e+00}$ & $\mathbf{1.28e+01}$ & - & $3.44e+01$ & $2.78e+01$ & +\\
$f_{7}$ & \multicolumn{3}{c|}{$4.77e+01\,\pm\,1.90e+01$} & $\mathbf{7.02e+00}$ & $\mathbf{4.95e+00}$ & - & $3.51e+08$ & $3.49e+09$ & = & $1.19e+02$ & $2.33e+01$ & +\\
$f_{8}$ & \multicolumn{3}{c|}{$\mathbf{2.09e+01}\,\pm\,\mathbf{9.95e-02}$} & $2.10e+01$ & $5.69e-02$ & + & $2.10e+01$ & $5.49e-02$ & + & $2.10e+01$ & $5.44e-02$ & +\\
$f_{9}$ & \multicolumn{3}{c|}{$\mathbf{2.18e+01}\,\pm\,\mathbf{3.81e+00}$} & $3.37e+01$ & $1.73e+00$ & + & $4.42e+01$ & $7.09e+00$ & + & $3.02e+01$ & $2.20e+00$ & +\\
$f_{10}$ & \multicolumn{3}{c|}{$7.94e-02\,\pm\,3.82e-02$} & $4.30e-02$ & $2.36e-02$ & - & $\mathbf{2.01e-02}$ & $\mathbf{1.71e-02}$ & - & $2.00e-01$ & $9.45e-02$ & +\\
$f_{11}$ & \multicolumn{3}{c|}{$\mathbf{0.00e+00}\,\pm\,\mathbf{6.33e-14}$} & $3.79e+00$ & $3.63e+00$ & + & $1.05e+02$ & $2.55e+02$ & + & $5.76e-01$ & $6.49e-01$ & +\\
$f_{12}$ & \multicolumn{3}{c|}{$\mathbf{7.08e+01}\,\pm\,\mathbf{1.92e+01}$} & $1.39e+02$ & $1.53e+01$ & + & $8.08e+02$ & $9.37e+02$ & = & $2.13e+02$ & $5.62e+01$ & +\\
$f_{13}$ & \multicolumn{3}{c|}{$\mathbf{1.23e+02}\,\pm\,\mathbf{2.70e+01}$} & $1.56e+02$ & $1.42e+01$ & + & $1.65e+03$ & $1.67e+03$ & + & $2.58e+02$ & $4.39e+01$ & +\\
$f_{14}$ & \multicolumn{3}{c|}{$8.23e+01\,\pm\,4.30e+02$} & $1.22e+03$ & $1.96e+02$ & + & $5.39e+03$ & $7.64e+02$ & + & $\mathbf{6.57e+00}$ & $\mathbf{3.69e+00}$ & -\\
$f_{15}$ & \multicolumn{3}{c|}{$\mathbf{3.95e+03}\,\pm\,\mathbf{6.67e+02}$} & $6.92e+03$ & $3.27e+02$ & + & $5.29e+03$ & $6.36e+02$ & + & $4.03e+03$ & $4.77e+02$ & =\\
$f_{16}$ & \multicolumn{3}{c|}{$5.89e-01\,\pm\,2.84e-01$} & $2.58e+00$ & $2.96e-01$ & + & $\mathbf{1.23e-01}$ & $\mathbf{1.06e-01}$ & - & $2.40e+00$ & $4.03e-01$ & +\\
$f_{17}$ & \multicolumn{3}{c|}{$3.20e+01\,\pm\,4.84e-01$} & $6.02e+01$ & $3.72e+00$ & + & $4.07e+03$ & $8.51e+02$ & + & $\mathbf{3.13e+01}$ & $\mathbf{4.89e-01}$ & -\\
$f_{18}$ & \multicolumn{3}{c|}{$\mathbf{8.75e+01}\,\pm\,\mathbf{2.06e+01}$} & $2.04e+02$ & $1.15e+01$ & + & $3.95e+03$ & $7.79e+02$ & + & $2.44e+02$ & $5.78e+01$ & +\\
$f_{19}$ & \multicolumn{3}{c|}{$1.58e+00\,\pm\,3.50e-01$} & $5.54e+00$ & $5.23e-01$ & + & $3.50e+00$ & $9.05e-01$ & + & $\mathbf{8.55e-01}$ & $\mathbf{1.71e-01}$ & -\\
$f_{20}$ & \multicolumn{3}{c|}{$1.39e+01\,\pm\,1.29e+00$} & $\mathbf{1.24e+01}$ & $\mathbf{2.88e-01}$ & - & $1.50e+01$ & $4.97e-02$ & + & $1.39e+01$ & $4.52e-01$ & -\\
$f_{21}$ & \multicolumn{3}{c|}{$3.11e+02\,\pm\,8.08e+01$} & $2.92e+02$ & $7.29e+01$ & - & $3.09e+02$ & $8.58e+01$ & + & $\mathbf{2.58e+02}$ & $\mathbf{7.21e+01}$ & =\\
$f_{22}$ & \multicolumn{3}{c|}{$1.66e+02\,\pm\,2.19e+02$} & $2.03e+03$ & $2.77e+02$ & + & $6.92e+03$ & $9.35e+02$ & + & $\mathbf{1.21e+02}$ & $\mathbf{7.28e+01}$ & -\\
$f_{23}$ & \multicolumn{3}{c|}{$\mathbf{4.74e+03}\,\pm\,\mathbf{7.02e+02}$} & $7.30e+03$ & $3.62e+02$ & + & $6.78e+03$ & $7.36e+02$ & + & $5.26e+03$ & $7.22e+02$ & +\\
$f_{24}$ & \multicolumn{3}{c|}{$2.26e+02\,\pm\,9.00e+00$} & $\mathbf{2.07e+02}$ & $\mathbf{6.88e+00}$ & - & $7.93e+02$ & $5.89e+02$ & + & $2.81e+02$ & $1.08e+01$ & +\\
$f_{25}$ & \multicolumn{3}{c|}{$\mathbf{2.82e+02}\,\pm\,\mathbf{1.07e+01}$} & $2.89e+02$ & $1.65e+01$ & + & $3.81e+02$ & $1.54e+02$ & + & $3.03e+02$ & $6.25e+00$ & +\\
$f_{26}$ & \multicolumn{3}{c|}{$2.03e+02\,\pm\,2.21e+01$} & $\mathbf{2.00e+02}$ & $\mathbf{5.63e-02}$ & + & $4.66e+02$ & $4.25e+02$ & + & $2.02e+02$ & $4.53e+00$ & +\\
$f_{27}$ & \multicolumn{3}{c|}{$8.20e+02\,\pm\,1.48e+02$} & $\mathbf{7.91e+02}$ & $\mathbf{2.83e+02}$ & = & $8.17e+02$ & $2.09e+02$ & = & $1.07e+03$ & $1.13e+02$ & +\\
$f_{28}$ & \multicolumn{3}{c|}{$3.08e+02\,\pm\,1.06e+02$} & $\mathbf{3.00e+02}$ & $\mathbf{9.30e-13}$ & - & $1.94e+03$ & $3.38e+03$ & + & $5.43e+02$ & $5.77e+02$ & +\\
\hline\hline
\end{tabular}
\end{tiny}
\end{center}
\end{table*}

\begin{table*}[!ht]
\begin{center}
\captionsetup{justification=centering,width=22cm,margin=0pt,margin=0pt}
\caption{\label{tab:res50a} Statistical comparison of MS-CAP against jDE, CMA-ES, and CCPSO2 on CEC 2013 in $50$ dimensions}
\vspace{-0.3cm}
\begin{tiny}
\begin{tabular}{l|r@{$\,\pm\,$}l|c|r@{$\,\pm\,$}l|c|r@{$\,\pm\,$}l|c|r@{$\,\pm\,$}l|c}
\hline\hline
              &      \multicolumn{3}{c|}{MS-CAP}    &   \multicolumn{3}{c|}{jDE}    &   \multicolumn{3}{c|}{CMA-ES}&   \multicolumn{3}{c}{CCPSO2}\\
\hline
$f_{1}$ & \multicolumn{3}{c|}{$2.27e-13\,\pm\,2.27e-14$} & $\mathbf{0.00e+00}$ & $\mathbf{2.19e-13}$ & - & $2.27e-13$ & $0.00e+00$ & = & $7.05e-12$ & $3.53e-11$ & +\\
$f_{2}$ & \multicolumn{3}{c|}{$9.35e+05\,\pm\,3.48e+05$} & $4.11e+06$ & $1.51e+06$ & + & $\mathbf{2.27e-13}$ & $\mathbf{0.00e+00}$ & - & $4.37e+06$ & $2.29e+06$ & +\\
$f_{3}$ & \multicolumn{3}{c|}{$4.38e+07\,\pm\,5.32e+07$} & $1.58e+07$ & $2.59e+07$ & - & $\mathbf{2.32e+04}$ & $\mathbf{9.57e+04}$ & - & $3.09e+09$ & $3.03e+09$ & +\\
$f_{4}$ & \multicolumn{3}{c|}{$3.37e+03\,\pm\,1.16e+03$} & $1.71e+04$ & $3.88e+03$ & + & $\mathbf{2.27e-13}$ & $\mathbf{0.00e+00}$ & - & $1.08e+05$ & $3.86e+04$ & +\\
$f_{5}$ & \multicolumn{3}{c|}{$1.14e-13\,\pm\,1.82e-13$} & $\mathbf{1.14e-13}$ & $\mathbf{4.82e-14}$ & - & $1.95e-09$ & $9.17e-10$ & + & $3.92e-04$ & $3.89e-03$ & +\\
$f_{6}$ & \multicolumn{3}{c|}{$4.62e+01\,\pm\,7.15e+00$} & $4.40e+01$ & $8.29e-01$ & - & $\mathbf{4.29e+01}$ & $\mathbf{5.98e+00}$ & - & $4.74e+01$ & $1.34e+01$ & -\\
$f_{7}$ & \multicolumn{3}{c|}{$7.42e+01\,\pm\,1.39e+01$} & $\mathbf{2.88e+01}$ & $\mathbf{1.08e+01}$ & - & $1.98e+04$ & $1.96e+05$ & - & $1.43e+02$ & $2.39e+01$ & +\\
$f_{8}$ & \multicolumn{3}{c|}{$\mathbf{2.11e+01}\,\pm\,\mathbf{6.64e-02}$} & $2.12e+01$ & $3.53e-02$ & + & $2.11e+01$ & $3.75e-02$ & + & $2.12e+01$ & $3.86e-02$ & +\\
$f_{9}$ & \multicolumn{3}{c|}{$\mathbf{4.65e+01}\,\pm\,\mathbf{6.28e+00}$} & $6.41e+01$ & $4.64e+00$ & + & $7.66e+01$ & $8.71e+00$ & + & $5.87e+01$ & $3.26e+00$ & +\\
$f_{10}$ & \multicolumn{3}{c|}{$1.40e-01\,\pm\,7.42e-02$} & $1.09e-01$ & $4.69e-02$ & - & $\mathbf{2.70e-02}$ & $\mathbf{1.55e-02}$ & - & $2.03e-01$ & $1.80e-01$ & +\\
$f_{11}$ & \multicolumn{3}{c|}{$\mathbf{9.95e-02}\,\pm\,\mathbf{2.98e-01}$} & $4.15e+01$ & $5.92e+00$ & + & $2.46e+02$ & $5.29e+02$ & + & $9.07e-01$ & $8.53e-01$ & +\\
$f_{12}$ & \multicolumn{3}{c|}{$\mathbf{1.56e+02}\,\pm\,\mathbf{3.65e+01}$} & $2.85e+02$ & $3.74e+01$ & + & $2.28e+03$ & $1.53e+03$ & + & $4.55e+02$ & $8.03e+01$ & +\\
$f_{13}$ & \multicolumn{3}{c|}{$\mathbf{2.90e+02}\,\pm\,\mathbf{5.36e+01}$} & $3.33e+02$ & $2.47e+01$ & + & $3.26e+03$ & $1.25e+03$ & + & $5.69e+02$ & $8.18e+01$ & +\\
$f_{14}$ & \multicolumn{3}{c|}{$1.09e+02\,\pm\,5.45e+02$} & $3.54e+03$ & $3.45e+02$ & + & $8.74e+03$ & $1.05e+03$ & + & $\mathbf{7.35e+00}$ & $\mathbf{3.55e+00}$ & -\\
$f_{15}$ & \multicolumn{3}{c|}{$\mathbf{7.29e+03}\,\pm\,\mathbf{8.50e+02}$} & $1.35e+04$ & $3.73e+02$ & + & $9.04e+03$ & $8.70e+02$ & + & $8.31e+03$ & $8.71e+02$ & +\\
$f_{16}$ & \multicolumn{3}{c|}{$9.92e-01\,\pm\,4.60e-01$} & $3.37e+00$ & $3.03e-01$ & + & $\mathbf{8.00e-02}$ & $\mathbf{4.27e-02}$ & - & $2.75e+00$ & $5.96e-01$ & +\\
$f_{17}$ & \multicolumn{3}{c|}{$5.37e+01\,\pm\,7.94e-01$} & $1.33e+02$ & $7.95e+00$ & + & $6.84e+03$ & $1.10e+03$ & + & $\mathbf{5.16e+01}$ & $\mathbf{3.28e-01}$ & -\\
$f_{18}$ & \multicolumn{3}{c|}{$\mathbf{1.66e+02}\,\pm\,\mathbf{3.39e+01}$} & $3.99e+02$ & $1.75e+01$ & + & $7.01e+03$ & $9.83e+02$ & + & $4.87e+02$ & $9.77e+01$ & +\\
$f_{19}$ & \multicolumn{3}{c|}{$2.71e+00\,\pm\,4.71e-01$} & $1.20e+01$ & $1.10e+00$ & + & $6.26e+00$ & $1.54e+00$ & + & $\mathbf{1.49e+00}$ & $\mathbf{2.32e-01}$ & -\\
$f_{20}$ & \multicolumn{3}{c|}{$2.34e+01\,\pm\,1.34e+00$} & $\mathbf{2.23e+01}$ & $\mathbf{3.31e-01}$ & - & $2.50e+01$ & $9.74e-02$ & + & $2.33e+01$ & $8.19e-01$ & -\\
$f_{21}$ & \multicolumn{3}{c|}{$8.43e+02\,\pm\,3.85e+02$} & $7.09e+02$ & $4.40e+02$ & = & $7.95e+02$ & $3.57e+02$ & = & $\mathbf{4.42e+02}$ & $\mathbf{3.45e+02}$ & -\\
$f_{22}$ & \multicolumn{3}{c|}{$2.13e+02\,\pm\,8.07e+02$} & $4.85e+03$ & $4.56e+02$ & + & $1.18e+04$ & $1.34e+03$ & + & $\mathbf{1.11e+02}$ & $\mathbf{9.60e+01}$ & -\\
$f_{23}$ & \multicolumn{3}{c|}{$\mathbf{8.90e+03}\,\pm\,\mathbf{1.14e+03}$} & $1.42e+04$ & $3.85e+02$ & + & $1.18e+04$ & $9.41e+02$ & + & $1.09e+04$ & $1.34e+03$ & +\\
$f_{24}$ & \multicolumn{3}{c|}{$2.69e+02\,\pm\,1.44e+01$} & $\mathbf{2.40e+02}$ & $\mathbf{1.65e+01}$ & - & $1.74e+03$ & $1.02e+03$ & + & $3.60e+02$ & $9.64e+00$ & +\\
$f_{25}$ & \multicolumn{3}{c|}{$\mathbf{3.64e+02}\,\pm\,\mathbf{1.74e+01}$} & $3.94e+02$ & $1.29e+01$ & + & $5.07e+02$ & $2.06e+02$ & + & $3.97e+02$ & $1.08e+01$ & +\\
$f_{26}$ & \multicolumn{3}{c|}{$2.27e+02\,\pm\,7.30e+01$} & $\mathbf{2.10e+02}$ & $\mathbf{4.54e+01}$ & + & $7.71e+02$ & $8.75e+02$ & + & $2.15e+02$ & $4.95e+01$ & +\\
$f_{27}$ & \multicolumn{3}{c|}{$1.44e+03\,\pm\,1.78e+02$} & $1.67e+03$ & $4.10e+02$ & + & $\mathbf{1.32e+03}$ & $\mathbf{3.23e+02}$ & - & $1.82e+03$ & $8.56e+01$ & +\\
$f_{28}$ & \multicolumn{3}{c|}{$4.00e+02\,\pm\,6.74e-13$} & $\mathbf{4.00e+02}$ & $\mathbf{4.36e-13}$ & - & $2.80e+03$ & $4.35e+03$ & + & $7.24e+02$ & $1.08e+03$ & +\\
\hline\hline
\end{tabular}
\end{tiny}
\end{center}
\end{table*}

\begin{table*}[!ht]
\begin{center}
\captionsetup{justification=centering,width=22cm,margin=0pt,margin=0pt}
\caption{\label{tab:res1000b} Statistical comparison of MS-CAP against jDE, CMA-ES, and CCPSO2 on CEC 2010 in $1000$ dimensions}
\vspace{-0.3cm}
\begin{tiny}
\begin{tabular}{l|r@{$\,\pm\,$}l|c|r@{$\,\pm\,$}l|c|r@{$\,\pm\,$}l|c|r@{$\,\pm\,$}l|c}
\hline\hline
              &      \multicolumn{3}{c|}{MS-CAP}    &   \multicolumn{3}{c|}{jDE}    &   \multicolumn{3}{c|}{CMA-ES}&   \multicolumn{3}{c}{CCPSO2}\\
\hline
$f_{1}$ & \multicolumn{3}{c|}{$5.13e-02\,\pm\,2.11e-01$} & $1.83e-07$ & $1.10e-06$ & - & $6.95e+04$ & $9.91e+03$ & + & $\mathbf{6.47e-14}$ & $\mathbf{1.41e-13}$ & -\\
$f_{2}$ & \multicolumn{3}{c|}{$9.64e+02\,\pm\,8.50e+02$} & $2.39e+03$ & $3.39e+02$ & + & $1.01e+04$ & $4.63e+02$ & + & $\mathbf{1.36e+02}$ & $\mathbf{1.11e+02}$ & -\\
$f_{3}$ & \multicolumn{3}{c|}{$1.19e+01\,\pm\,7.27e-01$} & $1.34e+01$ & $7.40e-01$ & + & $1.99e+01$ & $1.12e-02$ & + & $\mathbf{7.34e-11}$ & $\mathbf{1.05e-10}$ & -\\
$f_{4}$ & \multicolumn{3}{c|}{$8.69e+11\,\pm\,2.84e+11$} & $1.03e+12$ & $3.25e+11$ & + & $\mathbf{5.55e+10}$ & $\mathbf{4.75e+09}$ & - & $2.14e+12$ & $1.27e+12$ & +\\
$f_{5}$ & \multicolumn{3}{c|}{$1.18e+08\,\pm\,2.16e+07$} & $\mathbf{7.66e+07}$ & $\mathbf{1.58e+07}$ & - & $6.65e+08$ & $1.19e+08$ & + & $3.92e+08$ & $7.98e+07$ & +\\
$f_{6}$ & \multicolumn{3}{c|}{$1.81e+01\,\pm\,3.07e-01$} & $\mathbf{1.42e+01}$ & $\mathbf{7.06e-01}$ & - & $1.98e+07$ & $5.87e+04$ & + & $1.71e+07$ & $4.45e+06$ & +\\
$f_{7}$ & \multicolumn{3}{c|}{$1.77e+02\,\pm\,5.06e+02$} & $\mathbf{1.29e+01}$ & $\mathbf{3.68e+01}$ & - & $3.08e+06$ & $2.04e+05$ & + & $7.60e+09$ & $9.72e+09$ & +\\
$f_{8}$ & \multicolumn{3}{c|}{$3.26e+07\,\pm\,2.47e+07$} & $5.05e+07$ & $2.48e+07$ & + & $\mathbf{4.44e+06}$ & $\mathbf{3.21e+05}$ & - & $5.46e+07$ & $4.16e+07$ & +\\
$f_{9}$ & \multicolumn{3}{c|}{$7.65e+07\,\pm\,7.13e+06$} & $5.05e+07$ & $4.73e+06$ & - & $\mathbf{7.27e+04}$ & $\mathbf{1.07e+04}$ & - & $5.01e+07$ & $7.68e+06$ & -\\
$f_{10}$ & \multicolumn{3}{c|}{$4.52e+03\,\pm\,4.12e+02$} & $\mathbf{4.49e+03}$ & $\mathbf{9.53e+02}$ & = & $1.03e+04$ & $4.04e+02$ & + & $4.57e+03$ & $2.75e+02$ & +\\
$f_{11}$ & \multicolumn{3}{c|}{$1.83e+02\,\pm\,1.66e+01$} & $\mathbf{1.05e+02}$ & $\mathbf{1.55e+01}$ & - & $2.18e+02$ & $1.77e-01$ & + & $2.00e+02$ & $5.98e+00$ & +\\
$f_{12}$ & \multicolumn{3}{c|}{$1.17e+04\,\pm\,8.61e+02$} & $1.22e+06$ & $2.04e+06$ & + & $\mathbf{1.64e-19}$ & $\mathbf{4.18e-20}$ & - & $6.12e+04$ & $8.14e+04$ & +\\
$f_{13}$ & \multicolumn{3}{c|}{$1.49e+03\,\pm\,2.59e+02$} & $1.14e+03$ & $2.25e+02$ & - & $\mathbf{4.53e+01}$ & $\mathbf{6.59e+01}$ & - & $1.14e+03$ & $5.42e+02$ & -\\
$f_{14}$ & \multicolumn{3}{c|}{$2.69e+08\,\pm\,1.90e+07$} & $1.71e+08$ & $1.15e+07$ & - & $\mathbf{7.69e+04}$ & $\mathbf{1.06e+04}$ & - & $1.60e+08$ & $3.35e+07$ & -\\
$f_{15}$ & \multicolumn{3}{c|}{$7.59e+03\,\pm\,5.90e+02$} & $\mathbf{5.73e+03}$ & $\mathbf{1.22e+03}$ & - & $1.04e+04$ & $5.58e+02$ & + & $9.31e+03$ & $5.52e+02$ & +\\
$f_{16}$ & \multicolumn{3}{c|}{$3.86e+02\,\pm\,2.51e+00$} & $\mathbf{3.32e+02}$ & $\mathbf{2.43e+01}$ & - & $3.97e+02$ & $2.92e-01$ & + & $3.95e+02$ & $1.45e+00$ & +\\
$f_{17}$ & \multicolumn{3}{c|}{$8.44e+04\,\pm\,5.92e+03$} & $3.81e+06$ & $3.77e+06$ & + & $\mathbf{4.17e-19}$ & $\mathbf{7.23e-20}$ & - & $1.41e+05$ & $1.44e+05$ & +\\
$f_{18}$ & \multicolumn{3}{c|}{$4.81e+03\,\pm\,9.55e+02$} & $2.85e+03$ & $6.44e+02$ & - & $\mathbf{1.59e+02}$ & $\mathbf{1.67e+02}$ & - & $5.62e+03$ & $4.13e+03$ & =\\
$f_{19}$ & \multicolumn{3}{c|}{$7.02e+05\,\pm\,4.80e+04$} & $1.92e+07$ & $3.24e+06$ & + & $\mathbf{3.38e+01}$ & $\mathbf{1.36e+01}$ & - & $1.14e+06$ & $1.22e+06$ & =\\
$f_{20}$ & \multicolumn{3}{c|}{$3.21e+03\,\pm\,2.10e+02$} & $2.33e+03$ & $1.82e+02$ & - & $\mathbf{7.51e+02}$ & $\mathbf{9.99e+01}$ & - & $1.42e+03$ & $1.19e+02$ & -\\
\hline\hline
\end{tabular}
\end{tiny}
\end{center}
\end{table*}


A summary of the statistical comparisons, against all the ten algorithms under examination, is presented in Table~\ref{tab:stat-summary}. In total, $1040$ pairwise comparisons were performed ($10$ algorithms, each one tested on $104$ test problems, i.e. $28$ CEC 2013 test functions in $10$, $30$ and $50$ dimensions plus $20$ CEC 2010 test functions in $1000$ dimensions). 

\begin{table*}[!ht]
\begin{center}
\captionsetup{justification=centering,width=22cm,margin=0pt,margin=0pt}
\caption{\label{tab:stat-summary}Summary of the pairwise statistical comparisons}
\begin{tiny}
\begin{tabular}{c|c|c|c|c|c}
\hline\hline
\multirow{3}{*}{Optimizer} & \multicolumn{4}{c|}{Problem dimension} & \multirow{2}{*}{TOT}\\
\cline{2-5}
 & 10 & 30 & 50 & 1000 & \\
 & (-/=/+) & (-/=/+) & (-/=/+) & (-/=/+) & (-/=/+)\\
\hline
SADE & 6/5/17 & 6/4/18 & 5/2/21 & 3/0/17 & 20/11/73\\
JADE & 1/8/19 & 4/2/22 & 3/3/22 & 1/1/18 & 9/14/81\\
jDE & 5/3/20 & 10/1/17 & 9/1/18 & 12/1/7 & 36/6/62\\
MDE-pBX & 8/8/12 & 7/4/17 & 5/3/20 & 1/1/18 & 21/16/67\\
EPSDE & 2/4/22 & 9/0/19 & 8/1/19 & 9/1/10 & 28/6/70\\
CLPSO & 2/2/24 & 1/0/27 & 1/0/27 & 4/0/16 & 8/2/94\\
CCPSO2 & 0/0/28 & 5/2/21 & 7/0/21 & 7/2/11 & 19/4/81\\
PMS & 0/1/27 & 2/2/24 & 4/0/24 & 12/2/6 & 18/5/81\\
MACh & 9/2/17 & 8/2/18 & 9/2/17 & 0/0/20 & 26/6/72\\
CMA-ES & 5/1/22 & 6/4/18 & 8/2/18 & 10/0/10 & 29/7/68\\
\hline
TOT (-/=/+) & 38/34/208 & 58/21/201 & 59/14/207 & 59/8/133 & 214/77/749\\
\hline\hline
\end{tabular}
\end{tiny}
\end{center}
\end{table*}

\begin{figure*}
\includegraphics[width=0.5\columnwidth]{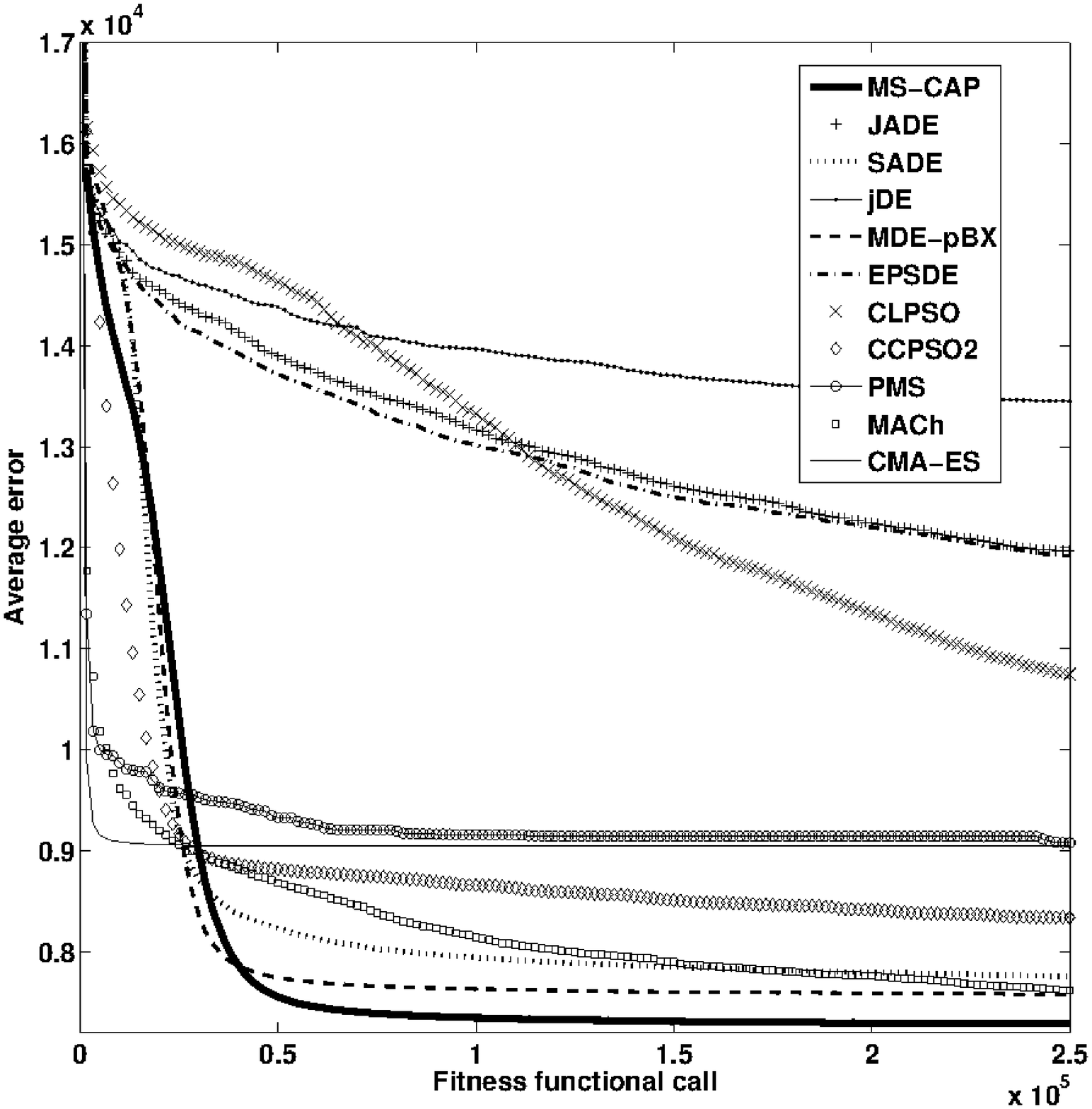}
\caption{Average fitness trend of the algorithms on test function $f_{15}$ from CEC 2013 in $50$ dimensions}\label{fig:res-cec13-f15-50}
\end{figure*}

It can be seen that MS-CAP is superior to the other algorithms in $749$ experiments ($72\%$), while it is outperformed in $20.58\%$ of cases. Looking at the aggregate pairwise algorithm comparisons, it emerges that MS-CAP outperforms all the other algorithms in most of the cases. The overall superior performance of MS-CAP is particularly evident against CLPSO, CCPSO2 and PMS, while jDE results the second most competitive algorithm after MS-CAP. Another interesting observation is that the performance of MS-CAP is very good at all the dimensionalities considered in our experiments, with a success rate of approximately $74\%$ in case of $10$, $30$ and $50$ dimensions, and $66.5\%$ in $1000$. 
Thus, although the performance of MS-CAP slightly deteriorates on large-scale problems (compared to jDE, EPSDE, PMS and CMA-ES), the proposed algorithm is able to provide competitive results, on diverse fitness landscapes, even in $1000$ dimensions. An example of fitness trend obtained during an optimization experiment is given in Fig.~\ref{fig:res-cec13-f15-50}, where it can be seen how, on that specific problem, MS-CAP converges much faster than the other algorithms towards the optimal solution. 

\subsection{Statistical Ranking through the Holm-Bonferroni Procedure}
\noindent
In order to give a further insight into the results presented above, we ranked the $11$ algorithms under study by means of the sequentially rejective Holm-Bonferroni procedure\cite{bib:Holm1979}, as described in\cite{bib:Garcia2008b}. The procedure consists of the following. Considering the results obtained by all the algorithms on the two benchmarks (at four different dimensionalities), for each problem we assigned to each algorithm a score $R_i$ for $i = 1,\dots,N_A$ (where $N_A$ is the number of algorithms under analysis, $N_A = 11$ in our case), being $11$ the score of the algorithm displaying the best performance on that problem, $10$ the score of the second best, and so on. The algorithm displaying the worst performance scores $1$. These scores are then averaged, for each algorithm, over the whole set of test problems ($104$ in our case). The algorithms are sorted on the basis of these average scores. Indicating with $R_0$ the rank (i.e, the average score) of MS-CAP, taken as reference, and with $R_j$ for $j = 1,\dots,N_A-1$ the rank of the remaining 
$10$ algorithms, the values $z_j$ are calculated as
\begin{equation}
z_j = \frac{R_j - R_0}{\sqrt{\frac{N_A(N_A+1)}{6N_{TP}}}} 
\end{equation}
where $N_{TP}$ is the number of test problems in consideration ($N_{TP} = 104$ in our case). By means of the $z_j$ values, the corresponding cumulative normal distribution values $p_j$ are derived. These $p_j$ values are then compared to the corresponding $\delta /j$ where $\delta$ is the level of confidence, set to $0.05$: if $p_j < \delta /j$, the null-hypothesis (that MS-CAP has the same performance as the j-th algorithm) is rejected, otherwise is accepted as well as all the subsequent tests.

Table \ref{tab:holm-test} displays the ranks, $z_j$ values, $p_j$ values, and corresponding $\delta/j$ obtained in this way. The rank of MS-CAP is shown in parenthesis in the table caption. Moreover, we indicate whether the null-hypothesis is rejected or not. In this case, all the hypotheses are sequentially rejected. Moreover, it should be noted that the proposed MS-CAP has the highest average rank ($8.32$), while jDE ranks second ($6.64$). This result not only confirms the superiority of the proposed approach to the state-of-the-art algorithms under comparison, but also shows its applicability as an algorithm fit to tackle a broad range of optimization problems.

\begin{table*}[!ht]
\begin{center}
\caption{\label{tab:holm-test}Holm-Bonferroni procedure (reference: MS-CAP, Rank = 8.32e+00)}
\begin{tiny}
\begin{tabular}{c|c|c|c|c|c|c}
\hline\hline
$j$ & Optimizer & Rank & $z_j$ & $p_j$ & $\delta/j$ & Hypothesis\\
\hline
1 & jDE & 6.64e+00 & -3.98e+00 & 3.38e-05 & 5.00e-02 & Rejected\\
2 & SADE & 6.49e+00 & -4.35e+00 & 6.77e-06 & 2.50e-02 & Rejected\\
3 & EPSDE & 6.48e+00 & -4.37e+00 & 6.09e-06 & 1.67e-02 & Rejected\\
4 & MDE-pBX & 6.15e+00 & -5.15e+00 & 1.28e-07 & 1.25e-02 & Rejected\\
5 & CCPSO2 & 5.52e+00 & -6.66e+00 & 1.33e-11 & 1.00e-02 & Rejected\\
6 & JADE & 5.36e+00 & -7.05e+00 & 8.71e-13 & 8.33e-03 & Rejected\\
7 & CLPSO & 5.31e+00 & -7.17e+00 & 3.80e-13 & 7.14e-03 & Rejected\\
8 & PMS & 5.16e+00 & -7.51e+00 & 2.92e-14 & 6.25e-03 & Rejected\\
9 & CMA-ES & 5.08e+00 & -7.72e+00 & 5.92e-15 & 5.56e-03 & Rejected\\
10 & MACh & 4.78e+00 & -8.43e+00 & 1.76e-17 & 5.00e-03 & Rejected\\
\hline\hline
\end{tabular}
\end{tiny}
\end{center}
\end{table*}

\subsection{Sensitivity of the Parameters}
\noindent
We performed an analysis of sensitivity varying, independently, the values of $N$ and $\varepsilon$. We replicated the experiments on the entire CEC 2013 testbed in $10$ and $30$ dimensions, and performed the statistical analysis as described before, using the results obtained with $N=50$ and $\varepsilon=10^{-6}$ as reference. Numerical results in $10$ dimensions are given in Tables \ref{tab:cec2013paramtun1} and \ref{tab:cec2013paramtun2}, where for each function the variation of performance depending on $N$ and $\varepsilon$, respectively, is shown. Also in this case, $100$ repetitions per function are considered. The results in $30$ dimensions, not reported here for brevity, are available at the link \url{https://sites.google.com/site/facaraff/home/Downloads/MS-CAP_Detailed_Results.pdf}. As shown MS-CAP appears to be fairly robust and its performance does not seem to be very sensitive to parameter variations, especially with respect to the decay threshold $\varepsilon$. As for the population size $N_p$, it seems that its influence on the algorithmic performance is slightly stronger. All in all, the parameter setting ($N=50$, $\varepsilon=10^{-6}$) provides the best results, guaranteeing the best trade-off in terms of optimization and scalability.

\begin{table*}
\caption{Average error $\pm$ standard deviation and statistical comparison (reference: MS-CAP $N=50$ $\varepsilon=10^{-6}$) for MS-CAP parameter tuning of $N$ on CEC 2013 in 10 dimensions}\label{tab:cec2013paramtun1}
\begin{center}
\begin{tiny}
\begin{tabular}{l|r@{$\,\pm\,$}l|c|r@{$\,\pm\,$}l|c|r@{$\,\pm\,$}l|c|r@{$\,\pm\,$}l|c}
\hline\hline
    &         \multicolumn{3}{c|}{$N=50$ $\varepsilon=10^{-6}$}    &   \multicolumn{3}{c|}{$N=10$ $\varepsilon=10^{-6}$}    &   \multicolumn{3}{c|}{$N=30$ $\varepsilon=10^{-6}$}    &   \multicolumn{3}{c}{$N=100$ $\varepsilon=10^{-6}$}    \\
              \hline
$f_{1}$ & \multicolumn{3}{c|}{$0.000e+00\,\pm\,0.000e+00$} & ${8.527e-14}$ & ${1.101e-13}$ & = & ${0.000e+00}$ & ${0.000e+00}$ & = & ${1.160e-11}$ & ${3.361e-12}$ & +   \\
$f_{2}$ & \multicolumn{3}{c|}{$3.711e+02\,\pm\,3.244e+02$} & ${1.849e+04}$ & ${1.835e+04}$ & + & ${7.401e+03}$ & ${7.480e+03}$ & + & ${3.084e+03}$ & ${1.081e+03}$ & +  \\
$f_{3}$ & \multicolumn{3}{c|}{$1.902e+02\,\pm\,4.859e+02$} & ${5.383e+07}$ & ${9.522e+07}$ & + & ${5.318e+02}$ & ${1.391e+03}$ & = & ${8.262e+05}$ & ${2.172e+06}$ & +  \\
$f_{4}$ & \multicolumn{3}{c|}{$5.948e+00\,\pm\,6.711e+00$} & ${1.286e+02}$ & ${1.218e+02}$ & + & ${1.263e+01}$ & ${1.993e+01}$ & = & ${2.759e+01}$ & ${8.841e+00}$ & + \\
$f_{5}$ & \multicolumn{3}{c|}{$0.000e+00\,\pm\,0.000e+00$} & ${8.527e-14}$ & ${7.520e-14}$ & + & ${0.000e+00}$ & ${0.000e+00}$ & = & ${3.688e-08}$ & ${1.508e-08}$ & +   \\
$f_{6}$ & \multicolumn{3}{c|}{$3.680e+00\,\pm\,4.750e+00$} & ${4.970e+00}$ & ${4.843e+00}$ & = & ${7.359e+00}$ & ${4.249e+00}$ & = & ${7.362e+00}$ & ${4.245e+00}$ & + \\
$f_{7}$ & \multicolumn{3}{c|}{$4.571e-01\,\pm\,8.440e-01$} & ${2.666e+01}$ & ${1.856e+01}$ & + & ${5.133e+00}$ & ${7.999e+00}$ & = & ${2.623e-01}$ & ${9.202e-02}$ & =  \\
$f_{8}$ & \multicolumn{3}{c|}{$2.029e+01\,\pm\,7.628e-02$} & ${2.047e+01}$ & ${1.330e-01}$ & + & ${2.028e+01}$ & ${8.395e-02}$ & = & ${2.030e+01}$ & ${1.224e-01}$ & =  \\
$f_{9}$ & \multicolumn{3}{c|}{$3.412e+00\,\pm\,1.443e+00$} & ${3.598e+00}$ & ${7.777e-01}$ & = & ${2.985e+00}$ & ${2.093e+00}$ & = & ${2.810e+00}$ & ${1.061e+00}$ & =   \\
$f_{10}$ & \multicolumn{3}{c|}{$6.217e-02\,\pm\,4.675e-02$} & ${4.597e-01}$ & ${4.803e-01}$ & + & ${1.495e-01}$ & ${8.180e-02}$ & + & ${8.365e-02}$ & ${5.842e-02}$ & =   \\
$f_{11}$ & \multicolumn{3}{c|}{$0.000e+00\,\pm\,0.000e+00$} & ${1.119e+00}$ & ${1.607e+00}$ & + & ${0.000e+00}$ & ${0.000e+00}$ & = & ${3.887e-02}$ & ${3.902e-02}$ & +\\
$f_{12}$ & \multicolumn{3}{c|}{$1.082e+01\,\pm\,4.986e+00$} & ${1.555e+01}$ & ${7.310e+00}$ & = & ${9.825e+00}$ & ${2.459e+00}$ & = & ${1.306e+01}$ & ${3.931e+00}$ & =  \\
$f_{13}$ & \multicolumn{3}{c|}{$1.592e+01\,\pm\,7.773e+00$} & ${2.966e+01}$ & ${9.341e+00}$ & + & ${1.423e+01}$ & ${6.185e+00}$ & = & ${1.479e+01}$ & ${6.203e+00}$ & = \\
$f_{14}$ & \multicolumn{3}{c|}{$6.678e+01\,\pm\,1.164e+02$} & ${2.158e+02}$ & ${4.326e+02}$ & = & ${9.630e-01}$ & ${1.417e+00}$ & - & ${9.764e+01}$ & ${7.797e+01}$ & =  \\
$f_{15}$ & \multicolumn{3}{c|}{$7.929e+02\,\pm\,2.454e+02$} & ${8.842e+02}$ & ${1.589e+02}$ & = & ${8.605e+02}$ & ${3.105e+02}$ & = & ${7.640e+02}$ & ${1.673e+02}$ & =   \\
$f_{16}$ & \multicolumn{3}{c|}{$2.459e-01\,\pm\,1.828e-01$} & ${3.139e-01}$ & ${1.849e-01}$ & = & ${3.001e-01}$ & ${2.394e-01}$ & = & ${3.555e-01}$ & ${1.563e-01}$ & =  \\
$f_{17}$ & \multicolumn{3}{c|}{$1.034e+01\,\pm\,1.172e-01$} & ${1.056e+01}$ & ${3.581e-01}$ & = & ${1.015e+01}$ & ${5.831e-02}$ & - & ${1.315e+01}$ & ${8.952e-01}$ & + \\
$f_{18}$ & \multicolumn{3}{c|}{$1.996e+01\,\pm\,5.780e+00$} & ${2.311e+01}$ & ${2.883e+00}$ & = & ${2.008e+01}$ & ${5.247e+00}$ & = & ${1.806e+01}$ & ${2.657e+00}$ & =\\
$f_{19}$ & \multicolumn{3}{c|}{$3.840e-01\,\pm\,1.458e-01$} & ${4.533e-01}$ & ${1.454e-01}$ & = & ${4.215e-01}$ & ${1.520e-01}$ & = & ${6.656e-01}$ & ${1.591e-01}$ & +  \\
$f_{20}$ & \multicolumn{3}{c|}{$2.685e+00\,\pm\,9.982e-01$} & ${3.236e+00}$ & ${4.941e-01}$ & = & ${3.261e+00}$ & ${4.124e-01}$ & = & ${2.610e+00}$ & ${3.668e-01}$ & = \\
$f_{21}$ & \multicolumn{3}{c|}{$4.002e+02\,\pm\,0.000e+00$} & ${4.002e+02}$ & ${8.039e-14}$ & = & ${4.002e+02}$ & ${0.000e+00}$ & = & ${4.002e+02}$ & ${3.591e-12}$ & +  \\
$f_{22}$ & \multicolumn{3}{c|}{$7.069e+01\,\pm\,6.495e+01$} & ${1.616e+02}$ & ${1.227e+02}$ & = & ${2.228e+01}$ & ${1.307e+01}$ & = & ${2.172e+02}$ & ${9.220e+01}$ & +  \\
$f_{23}$ & \multicolumn{3}{c|}{$9.519e+02\,\pm\,3.095e+02$} & ${1.301e+03}$ & ${3.464e+02}$ & = & ${9.606e+02}$ & ${2.439e+02}$ & = & ${9.099e+02}$ & ${3.228e+02}$ & =  \\
$f_{24}$ & \multicolumn{3}{c|}{$1.740e+02\,\pm\,3.460e+01$} & ${1.826e+02}$ & ${3.136e+01}$ & = & ${1.839e+02}$ & ${3.865e+01}$ & = & ${1.530e+02}$ & ${4.010e+01}$ & =   \\
$f_{25}$ & \multicolumn{3}{c|}{$2.012e+02\,\pm\,2.585e+00$} & ${2.088e+02}$ & ${3.972e+00}$ & + & ${1.911e+02}$ & ${3.018e+01}$ & = & ${2.024e+02}$ & ${3.109e+00}$ & =  \\
$f_{26}$ & \multicolumn{3}{c|}{$1.274e+02\,\pm\,2.874e+01$} & ${1.404e+02}$ & ${3.619e+01}$ & = & ${1.174e+02}$ & ${1.208e+01}$ & = & ${1.436e+02}$ & ${4.381e+01}$ & = \\
$f_{27}$ & \multicolumn{3}{c|}{$3.126e+02\,\pm\,3.303e+01$} & ${3.191e+02}$ & ${3.089e+01}$ & + & ${3.125e+02}$ & ${3.305e+01}$ & = & ${3.381e+02}$ & ${4.794e+01}$ & + \\
$f_{28}$ & \multicolumn{3}{c|}{$2.500e+02\,\pm\,8.660e+01$} & ${2.750e+02}$ & ${6.614e+01}$ & = & ${2.750e+02}$ & ${6.614e+01}$ & = & ${2.750e+02}$ & ${6.614e+01}$ & + \\
\hline\hline
\end{tabular}
\end{tiny}
\end{center}
\end{table*}

\begin{table*}
\caption{Average error $\pm$ standard deviation and statistical comparison (reference: MS-CAP $N=50$ $\varepsilon=10^{-6}$) for MS-CAP parameter tuning of $\varepsilon$ on CEC 2013 in 10 dimensions}\label{tab:cec2013paramtun2}
\begin{center}
\begin{tiny}
\begin{tabular}{l|r@{$\,\pm\,$}l|c|r@{$\,\pm\,$}l|c|r@{$\,\pm\,$}l|c|r@{$\,\pm\,$}l|c}
\hline\hline 
 & \multicolumn{3}{c|}{$N=50$ $\varepsilon=10^{-1}$} &   \multicolumn{3}{c|}{$N=50$ $\varepsilon=10^{-3}$} & \multicolumn{3}{c|}{$N=50$ $\varepsilon=10^{-5}$} & \multicolumn{3}{c}{$N=50$ $\varepsilon=10^{-7}$} \\
\hline
$f_{1}$ &${0.000e+00}$ & ${0.000e+00}$ & = & ${0.000e+00}$ & ${0.000e+00}$ & = & ${0.000e+00}$ & ${0.000e+00}$ & = & ${0.000e+00}$ & ${0.000e+00}$ & = \\
$f_{2}$ &${2.226e+04}$ & ${2.411e+04}$ & + & ${4.965e+03}$ & ${7.505e+03}$ & + & ${3.075e+03}$ & ${3.096e+03}$ & = & ${2.851e+02}$ & ${2.992e+02}$ & = \\
$f_{3}$ &${1.327e+02}$ & ${3.426e+02}$ & = & ${1.486e+02}$ & ${3.704e+02}$ & = & ${4.441e+01}$ & ${7.175e+01}$ & = & ${6.706e+01}$ & ${1.532e+02}$ & = \\
$f_{4}$ &${1.290e+02}$ & ${1.611e+02}$ & + & ${1.383e+01}$ & ${1.522e+01}$ & = & ${3.934e+00}$ & ${3.584e+00}$ & = & ${2.729e+00}$ & ${3.859e+00}$ & = \\
$f_{5}$ &${0.000e+00}$ & ${0.000e+00}$ & = & ${0.000e+00}$ & ${0.000e+00}$ & = & ${0.000e+00}$ & ${0.000e+00}$ & = & ${0.000e+00}$ & ${0.000e+00}$ & = \\
$f_{6}$ &${7.359e+00}$ & ${4.249e+00}$ & = & ${1.227e+00}$ & ${3.245e+00}$ & = & ${6.133e+00}$ & ${4.750e+00}$ & = & ${6.133e+00}$ & ${4.750e+00}$ & + \\
$f_{7}$ &${1.775e-01}$ & ${1.751e-01}$ & = & ${1.039e+00}$ & ${1.983e+00}$ & = & ${1.478e+00}$ & ${3.198e+00}$ & = & ${3.016e-01}$ & ${3.929e-01}$ & = \\
$f_{8}$ &${2.054e+01}$ & ${1.142e-01}$ & + & ${2.044e+01}$ & ${8.397e-02}$ & + & ${2.035e+01}$ & ${1.762e-01}$ & = & ${2.037e+01}$ & ${7.057e-02}$ & = \\
$f_{9}$ &${1.981e+00}$ & ${1.197e+00}$ & = & ${2.241e+00}$ & ${1.488e+00}$ & = & ${1.983e+00}$ & ${1.109e+00}$ & = & ${3.337e+00}$ & ${1.435e+00}$ & = \\
$f_{10}$ &${3.320e-02}$ & ${1.742e-02}$ & = & ${9.625e-02}$ & ${5.286e-02}$ & = & ${1.104e-01}$ & ${5.139e-02}$ & = & ${8.643e-02}$ & ${5.478e-02}$ & = \\
$f_{11}$ & ${1.244e-01}$ & ${3.291e-01}$ & = & ${0.000e+00}$ & ${0.000e+00}$ & = & ${0.000e+00}$ & ${0.000e+00}$ & = & ${0.000e+00}$ & ${0.000e+00}$ & = \\
$f_{12}$ & ${1.070e+01}$ & ${5.306e+00}$ & = & ${1.505e+01}$ & ${7.904e+00}$ & = & ${1.119e+01}$ & ${4.498e+00}$ & = & ${1.057e+01}$ & ${4.249e+00}$ & = \\
$f_{13}$ & ${1.414e+01}$ & ${7.612e+00}$ & = & ${1.621e+01}$ & ${8.909e+00}$ & = & ${2.015e+01}$ & ${7.202e+00}$ & = & ${1.141e+01}$ & ${3.922e+00}$ & = \\
$f_{14}$ & ${1.795e+01}$ & ${1.928e+01}$ & = & ${4.360e+00}$ & ${5.505e+00}$ & = & ${2.615e+00}$ & ${2.698e+00}$ & = & ${4.942e+00}$ & ${3.925e+00}$ & = \\
$f_{15}$ & ${9.289e+02}$ & ${2.976e+02}$ & = & ${8.842e+02}$ & ${2.197e+02}$ & = & ${8.129e+02}$ & ${1.965e+02}$ & = & ${6.786e+02}$ & ${2.204e+02}$ & = \\
$f_{16}$ & ${8.496e-01}$ & ${4.547e-01}$ & + & ${3.159e-01}$ & ${2.571e-01}$ & = & ${1.607e-01}$ & ${1.713e-01}$ & = & ${3.161e-01}$ & ${2.835e-01}$ & =  \\
$f_{17}$ & ${1.054e+01}$ & ${4.400e-01}$ & = & ${1.024e+01}$ & ${6.812e-02}$ & = & ${1.037e+01}$ & ${1.428e-01}$ & = & ${1.048e+01}$ & ${1.705e-01}$ & = \\
$f_{18}$ &  ${2.223e+01}$ & ${7.151e+00}$ & = & ${2.390e+01}$ & ${5.943e+00}$ & = & ${2.507e+01}$ & ${5.033e+00}$ & + & ${2.432e+01}$ & ${4.199e+00}$ & =  \\
$f_{19}$ &   ${5.591e-01}$ & ${1.687e-01}$ & = & ${4.392e-01}$ & ${7.951e-02}$ & = & ${4.616e-01}$ & ${1.464e-01}$ & = & ${4.295e-01}$ & ${1.603e-01}$ & = \\
$f_{20}$ &${2.979e+00}$ & ${2.121e-01}$ & = & ${3.158e+00}$ & ${2.244e-01}$ & = & ${2.903e+00}$ & ${6.511e-01}$ & = & ${3.161e+00}$ & ${3.947e-01}$ & = \\
$f_{21}$ & ${3.752e+02}$ & ${6.621e+01}$ & = & ${4.002e+02}$ & ${0.000e+00}$ & = & ${3.752e+02}$ & ${6.621e+01}$ & = & ${4.002e+02}$ & ${0.000e+00}$ & = \\
$f_{22}$ & ${2.415e+02}$ & ${3.395e+02}$ & = & ${7.715e+01}$ & ${6.285e+01}$ & = & ${1.808e+02}$ & ${2.731e+02}$ & = & ${5.343e+01}$ & ${3.993e+01}$ & =\\
$f_{23}$ & ${1.025e+03}$ & ${2.524e+02}$ & = & ${9.207e+02}$ & ${3.829e+02}$ & = & ${8.845e+02}$ & ${3.416e+02}$ & = & ${8.588e+02}$ & ${2.931e+02}$ & = \\
$f_{24}$ & ${2.054e+02}$ & ${5.834e+00}$ & + & ${1.698e+02}$ & ${3.722e+01}$ & = & ${1.956e+02}$ & ${1.734e+01}$ & = & ${1.855e+02}$ & ${3.063e+01}$ & = \\
$f_{25}$ & ${2.032e+02}$ & ${5.413e+00}$ & = & ${1.999e+02}$ & ${8.584e+00}$ & = & ${2.024e+02}$ & ${4.351e+00}$ & = & ${1.803e+02}$ & ${3.465e+01}$ & =  \\
$f_{26}$ & ${1.787e+02}$ & ${3.690e+01}$ & + & ${1.492e+02}$ & ${3.948e+01}$ & = & ${1.243e+02}$ & ${2.889e+01}$ & = & ${1.361e+02}$ & ${3.722e+01}$ & = \\
$f_{27}$ & ${3.000e+02}$ & ${1.831e-02}$ & - & ${3.375e+02}$ & ${4.839e+01}$ & = & ${3.250e+02}$ & ${4.327e+01}$ & = & ${3.126e+02}$ & ${3.304e+01}$ & = \\
$f_{28}$ & ${3.000e+02}$ & ${3.487e-12}$ & = & ${3.000e+02}$ & ${3.507e-12}$ & = & ${3.000e+02}$ & ${7.354e-12}$ & = & ${2.750e+02}$ & ${6.614e+01}$ & =\\
 \hline \hline
\end{tabular}
\end{tiny}
\end{center}
\end{table*}

\section{Application to Neural Network Training}\label{sec:neural}
\noindent
To conclude the presentation of MS-CAP, we describe here an application in the context of industrial robotics. In particular, we consider as a case study the 
training of a model of the forward kinematics of an all-revolute robot arm.

Generally speaking, the forward kinematics of an all-revolute robot arm can be described as follows:
\begin{equation}\label{eq:forwKin}
 \mathbf{x} = f(\theta,\phi)
\end{equation}
where $\theta$ is the vector of joint (angular) positions, $\phi$ is the set of parameters describing the kinematic chain of the arm, and $\mathbf{f}(\cdot)$ is the homogeneous transformation matrix which translates the joint positions $\theta$ (in the joint space) into the configuration $\mathbf{x}$ (in the configuration space) of the end-effector of the arm. The end-effector configuration $\mathbf{x}$ is in general a 6-dimensional vector whose components are the Cartesian position $[x,y,z]$ of the end-effector, and its orientation described as a tern of Euler angles (e.g. roll, pitch, yaw). On the other hand, $\theta$ is a vector whose cardinality is the number of revolute joints. Finally, the structure of the parameter set $\phi$ depends on the adopted kinematic representation, usually based on the Denavit-Hartenberg convention~\cite{bib:denavit1955}.

\vspace*{20pt}
\includegraphics[width=0.5\columnwidth]{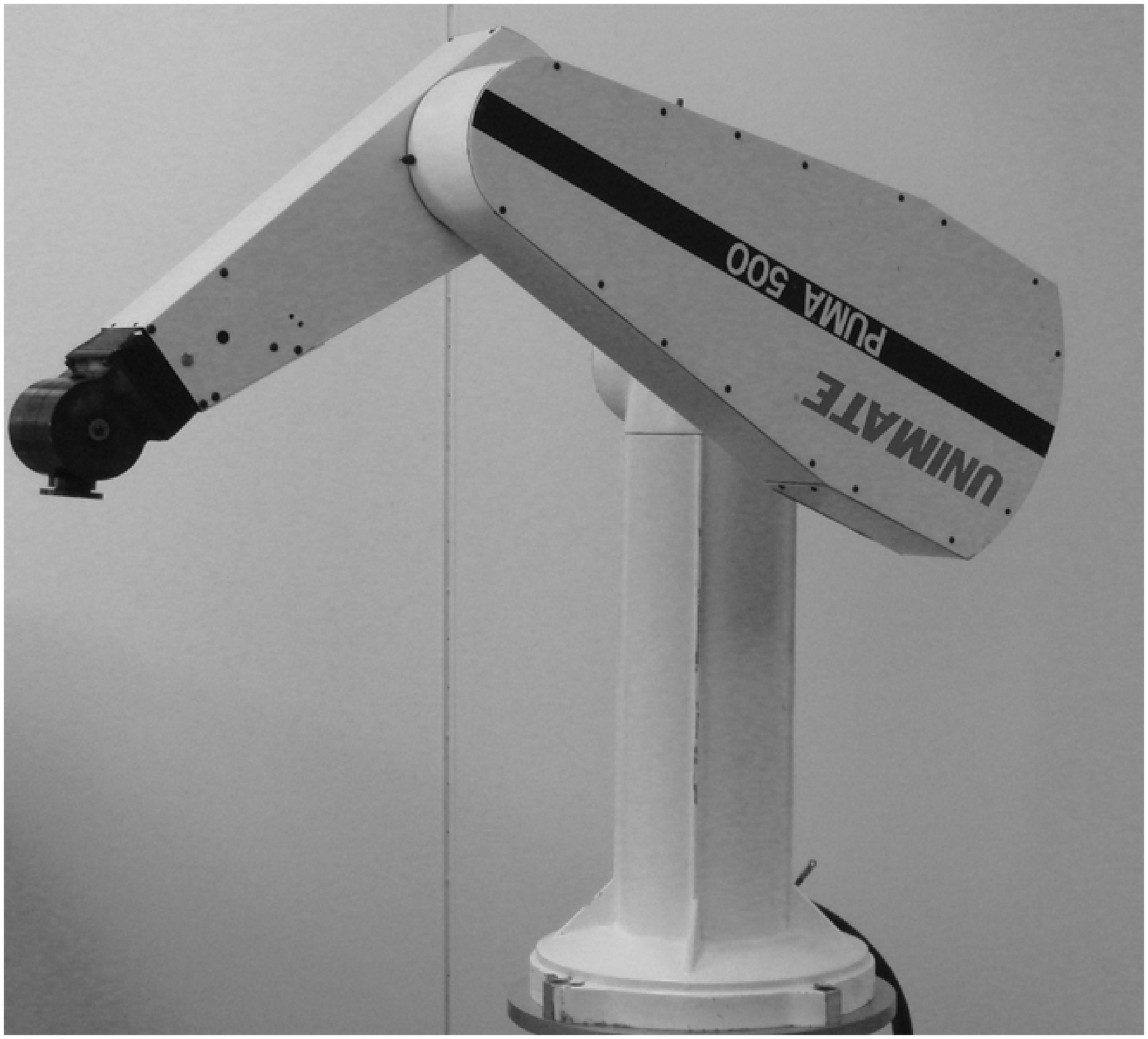}
\vspace*{-0.2cm}
\fcaption{\label{fig:robot-puma560}The PUMA 560 robotic arm}

For most robots the vector function $\mathbf{f}(\cdot)$ can be derived directly from the geometry of the manipulator. Nevertheless, when the structure of the robot is particularly complex, e.g. due to kinematic redundancy, writing this function is not trivial. In addition to that, the transformation matrix $\mathbf{f}(\cdot)$ is usually highly nonlinear and affected by noise, due e.g. manufacturing defects, wear of the robot components, and other factors of uncertainties. In all these cases, it is very important to have a reliable approximation of the forward kinematics model, especially for the purpose of control and path planning.

Following the problem definition described in~\cite{web:kinDatasets}, we consider here the task of predicting the Cartesian distance of the end-effector of an 8-link all-revolute manipulator from a predetermined target in the configuration space, given the angular positions of the eight joints. The target is arbitrarily set to the Cartesian position $[0.1, 0.1, 0.1]$ (relative distance, in meters, to the base frame of the robot arm). It is important to note that, compared to the general forward kinematics problem defined in eq.~(\ref{eq:forwKin}), here we are not interested in the whole configuration (position/orientation) of the end-effector, but only in its distance (thus, a single scalar value) from the target. The latter, however, obviously depending on the first. In this way, the kinematic model is a MISO model, i.e. with multiple inputs and a single output.

Among the eight different datasets available in~\cite{web:kinDatasets}, we consider two datasets. Both of them refer to the 8-dimensional, highly nonlinear case. The first is characterized by medium uniform noise while the second by high uniform noise. As reported in~\cite{web:kinDatasets}, the data are generated in Matlab using the Matlab Robotics Toolbox~\cite{bib:MatRotTool}, based on a realistic model of the 6-DOF PUMA 560 arm (Fig.~\ref{fig:robot-puma560}) with the addition of two fictitious joints to the end of the kinematic chain. The resulting datasets contain both $8192$ data points, each one consisting of eight angular positions and one distance value.

To model the kinematics of the robot, we consider a Feedforward Neural Network with eight input neurons (one per joint) and one output node. In the experiments, the neural network is implemented using the open-source Java package Encog~\cite{bib:encogBook}, version 3.2.0. To investigate the effect of the network architecture, we consider networks with $3$, $4$, and $5$ hidden nodes.
Each hidden node uses a sigmoid activation function with unitary slope.

The purpose of training the neural network consists in finding the optimal weights of the links in the network which guarantee the best approximation to the data. Indicating with $N_{hn}$ the number of hidden nodes, the total number of links is given by the number of input nodes ($8$ in our case) times $N_{hn}$, plus $N_{hn}$ links between the hidden nodes and the output node. Considering the three different hidden layer sizes, we then have optimization problems with $D=27$, $36$, and $45$ variables, respectively. 
The search space for each variable is $[-1,1]$. All the inputs/outputs are also normalized in the same range.

In total, we have six different optimization problems, corresponding to six combinations of datasets (medium and high noise) and hidden layer sizes ($3$, $4$, and $5$). For each problem, we divide the specific dataset in three equally sized subsets, respectively used for training, validating and testing the neural network. The training is performed minimizing the Mean Square Error (MSE) on the training subset. At the end of the training phase, the resulting neural network is then validated and tested on the two other subsets, and the corresponding MSE is calculated.

\subsection{Comparison against Meta-heuristics}
\noindent
In order to provide an exhaustive comparison, the MSE minimization process is performed using the same 11 algorithms presented in the previous section, with the same parameter setting. For each of the six configurations of dataset/hidden layer size, each algorithm has been executed $32$ independent times, each one continued for $5000~\times~D$ fitness evaluations.

\begin{figure*}
\includegraphics[width=0.5\columnwidth]{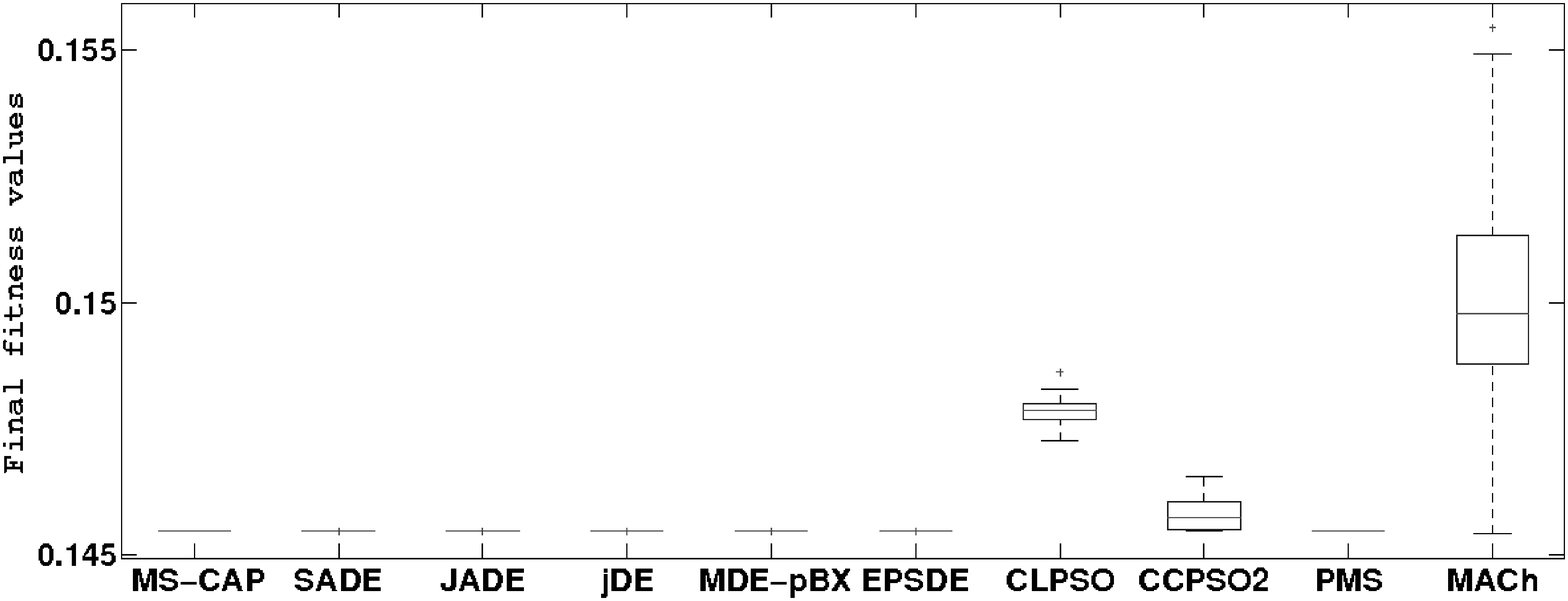}
\caption{Box plot of the final MSE values (over $32$ runs per algorithm) on the neural network training problem (high noise) for a neural network with four hidden nodes}\label{fig:robot-boxplot}
\end{figure*}

The experimental results in terms of average MSE (on the test subset) and standard deviation over $32$ runs at the end of the budget, as well as the statistical comparison as described in Section 3, are reported in Tables \ref{tab:NNmedium} and \ref{tab:NNhigh}, respectively for the medium and high noise cases. The box plot of the MSE values in the case of high noise and four hidden nodes is shown in Figure~\ref{fig:robot-boxplot} (the boxplots in the remaining five configurations are not shown because very similar). 

From the two tables and the figure, it can be seen that MS-CAP displays a respectable performance being as good as the best algorithms and clearly outperforming other competitors such as CLPSO, CCPSO2 and MACh. More specifically, the proposed algorithm statistically outperforms its competitors in $31$ cases, is outperformed in $23$ cases and displays a similar performance in $6$ cases. It can be noticed also that MS-CAP performs slightly better in the high noise case, and when the number of hidden nodes (which, in turn, affects the number of variables) increases. Interestingly, this result suggests on one hand that the MS-CAP algorithm is fairly robust against noise, on the other that it does not suffer from curse of dimensionality or overfitting.

\begin{figure*}
\includegraphics[width=0.5\columnwidth]{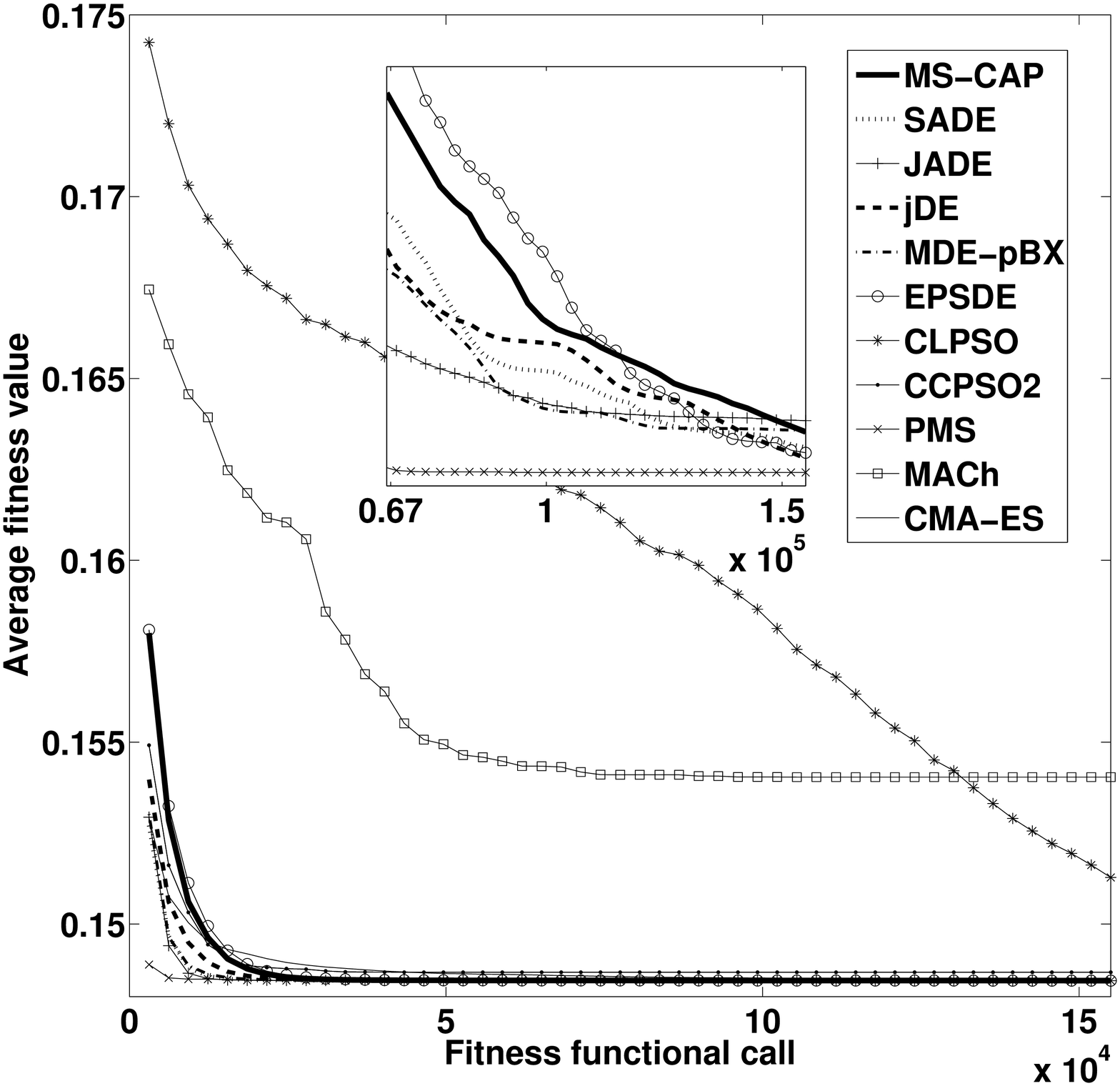}
\caption{Average MSE trend (over $32$ runs per algorithm) on the neural network training problem (medium noise) for a neural network with three hidden nodes}\label{fig:robot-trend1}
\end{figure*}

\begin{figure*}
\includegraphics[width=0.5\columnwidth]{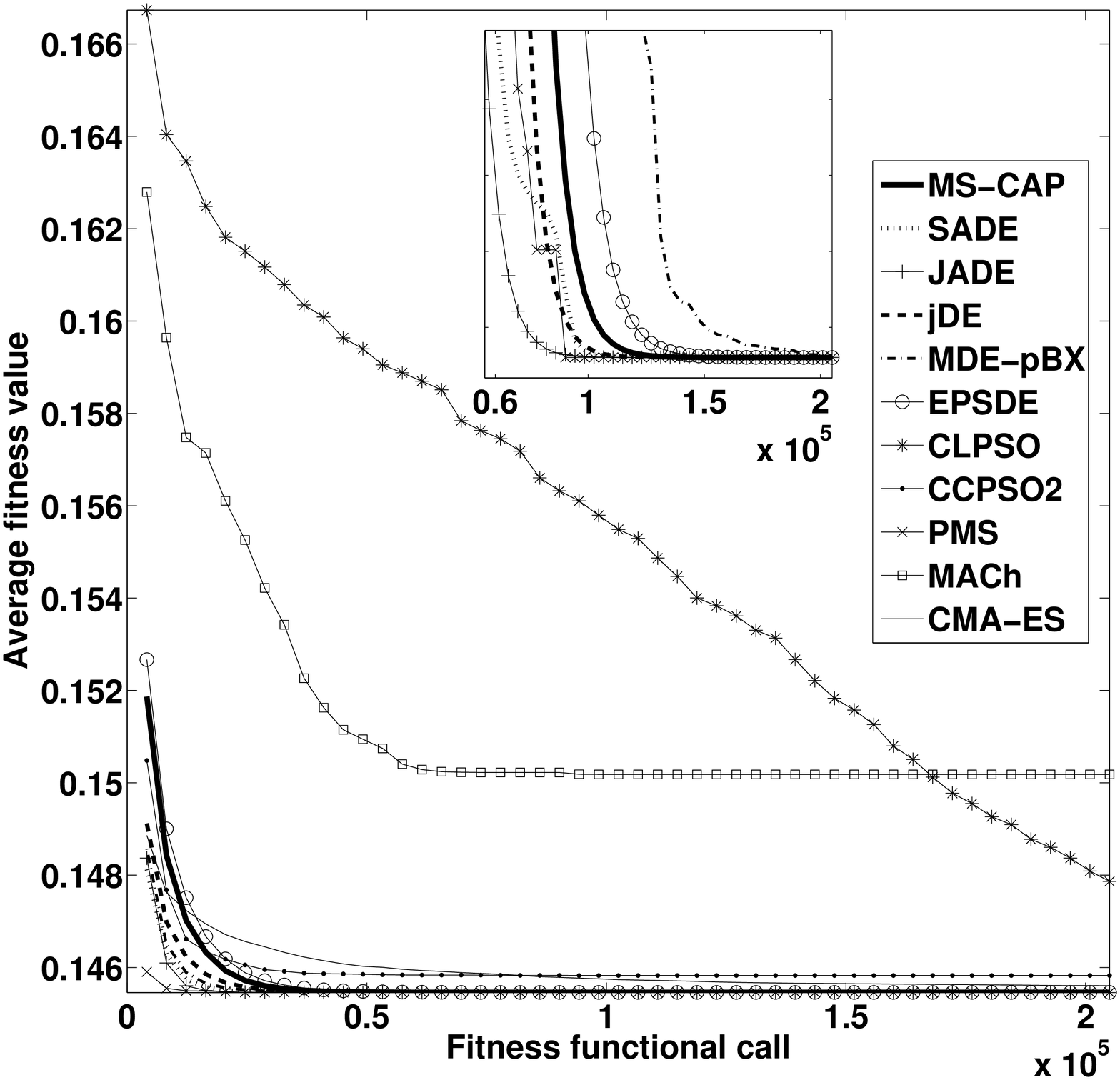}
\caption{Average MSE trend (over $32$ runs per algorithm) on the neural network training problem (high noise) for a neural network with four hidden nodes}\label{fig:robot-trend2}
\end{figure*}

The average MSE trends for two of the six configurations are shown in Figures~\ref{fig:robot-trend1} and \ref{fig:robot-trend2}. It can be observed that MS-CAP as well as several other competitors quickly detect solutions with a high quality while CLPSO and MACh display a much worse performance than the other algorithms (in particular, MACh suffers from premature convergence, while CLPSO converges slowly).

\begin{table*}
\caption{Average MSE $\pm$ standard deviation and statistical comparison (reference: MS-CAP) for MS-CAP against meta-heuristics on the neural network training problem (medium noise) for networks with $3$, $4$ and $5$ hidden nodes}\label{tab:NNmedium} 
\begin{center}
\begin{tiny}
\begin{tabular}{c|l@{$\,\pm\,$}r|c|l@{$\,\pm\,$}r|c|l@{$\,\pm\,$}r|c}
\hline\hline
Optimizer & \multicolumn{3}{c}{3 hidden nodes} & \multicolumn{3}{|c}{4 hidden nodes} & \multicolumn{3}{|c}{5 hidden nodes}\\
\hline
MS-CAP  &$1.48e-01$&$1.72e-06$&&$1.44e-01$&$8.16e-06$&&$1.40e-01$&$3.25e-06$&\\
SADE    &$1.48e-01$&$1.36e-06$&-&$1.44e-01$&$1.13e-05$&-&$1.40e-01$&$3.17e-06$&=\\
JADE    &$1.48e-01$&$2.01e-06$&-&$1.44e-01$&$2.53e-06$&-&$\mathbf{1.40e-01}$&$\mathbf{2.54e-06}$&-\\
jDE     &$1.48e-01$&$9.16e-07$&-&$\mathbf{1.44e-01}$&$\mathbf{2.74e-13}$&-&$1.40e-01$&$3.09e-06$&=\\
MDE-pBX &$1.48e-01$&$1.88e-06$&-&$1.44e-01$&$1.09e-05$&+&$1.40e-01$&$2.75e-06$&=\\
EPSDE   &$1.48e-01$&$1.23e-06$&=&$1.44e-01$&$3.14e-09$&+&$1.40e-01$&$3.23e-06$&=\\
CLPSO   &$1.51e-01$&$3.96e-04$&+&$1.47e-01$&$2.67e-04$&+&$1.43e-01$&$2.34e-04$&+\\
CCPSO2  &$1.49e-01$&$3.75e-04$&+&$1.45e-01$&$6.39e-04$&+&$1.41e-01$&$1.62e-04$&+\\
PMS     &$\mathbf{1.48e-01}$&$\mathbf{2.57e-10}$&-&$1.44e-01$&$1.89e-04$&-&$1.40e-01$&$2.62e-06$&-\\
MACh    &$1.54e-01$&$5.30e-03$&+&$1.50e-01$&$4.80e-03$&+&$1.49e-01$&$3.43e-03$&+\\
CMA-ES  &$1.48e-01$&$1.85e-05$&+&$1.44e-01$&$2.88e-05$&+&$1.41e-01$&$5.70e-05$&+\\
\hline\hline
\end{tabular}
\end{tiny}
\end{center}
\end{table*}

\begin{table*}
\caption{Average MSE $\pm$ standard deviation and statistical comparison (reference: MS-CAP) for MS-CAP against meta-heuristics on the neural network training problem (high noise) for networks with $3$, $4$ and $5$ hidden nodes}\label{tab:NNhigh}
\begin{center}
\begin{tiny}
\begin{tabular}{c|l@{$\,\pm\,$}r|c|l@{$\,\pm\,$}r|c|l@{$\,\pm\,$}r|c}
\hline\hline
Optimizer & \multicolumn{3}{c}{3 hidden nodes} & \multicolumn{3}{|c}{4 hidden nodes} & \multicolumn{3}{|c}{5 hidden nodes}\\
\hline
MS-CAP  &$1.50e-01$&$5.37e-15$&&$1.45e-01$&$5.59e-15$&&$1.44e-01$&$7.70e-15$&\\
SADE    &$1.50e-01$&$1.55e-15$&-&$1.45e-01$&$7.91e-14$&-&$1.44e-01$&$3.50e-12$&=\\
JADE    &$\mathbf{1.50e-01}$&$\mathbf{1.62e-16}$&-&$\mathbf{1.45e-01}$&$\mathbf{2.63e-16}$&-&$1.44e-01$&$3.71e-16$&-\\
jDE     &$1.50e-01$&$1.48e-16$&-&$1.45e-01$&$4.02e-16$&-&$\mathbf{1.44e-01}$&$\mathbf{3.17e-16}$&-\\
MDE-pBX &$1.50e-01$&$5.13e-10$&-&$1.45e-01$&$3.12e-09$&+&$1.44e-01$&$2.22e-09$&+\\
EPSDE   &$1.50e-01$&$5.71e-14$&+&$1.45e-01$&$2.86e-13$&+&$1.44e-01$&$1.27e-13$&+\\
CLPSO   &$1.53e-01$&$4.56e-04$&+&$1.48e-01$&$2.73e-04$&+&$1.46e-01$&$2.04e-04$&+\\
CCPSO2  &$1.51e-01$&$1.13e-03$&+&$1.46e-01$&$3.14e-04$&+&$1.44e-01$&$1.39e-04$&+\\
PMS     &$1.50e-01$&$1.01e-15$&-&$1.45e-01$&$9.13e-16$&-&$1.44e-01$&$9.76e-05$&-\\
MACh    &$1.56e-01$&$3.82e-03$&+&$1.50e-01$&$2.34e-03$&+&$1.48e-01$&$2.22e-03$&+\\
CMA-ES  &$1.50e-01$&$7.08e-05$&+&$1.46e-01$&$6.06e-05$&+&$1.44e-01$&$4.73e-05$&+\\
\hline\hline
\end{tabular}
\end{tiny}
\end{center}
\end{table*}

%
 
\subsection{Comparison against Classical Training Methods}
\noindent
We complete our study of the neural network training problem comparing the proposed MS-CAP against two classical training methods, namely the Error Back Propagation (EBP)\cite{bib:EBP} and the Resilient Propagation (RP)\cite{bib:RP}. For both the algorithms we used the implementation available in Encog, with the default parameter setting. In order to guarantee a fair comparison, we assign the same budget ($5000~\times~D$ evaluations) also to EBP and RP.

Numerical results related to the six configurations of dataset/hidden layer size defined before are reported in Tables \ref{tab:BACKNM} and \ref{tab:BACKNH}, respectively for the case of medium and high noise. It can be observed that, besides one case (3 hidden nodes, high noise), the proposed MS-CAP significantly outperforms the classical methods in terms of MSE. This result can be explained considering that the two classic algorithms, which are very specific to the training problem, perform better when the network size is smaller (in the presence of high noise); on the other hand, when the number of variables increases, a robust general-purpose optimizer tends to show a better performance. 

This experiment shows that MS-CAP is particularly suitable for training neural networks and more in general confirms, once again, that this efficient and versatile algorithm is able to obtain, with no prior tuning, a good performance on optimization problems from various domains.

\begin{table*}
\caption{Average MSE $\pm$ standard deviation and statistical comparison (reference: MS-CAP) for MS-CAP against EBP and RP on the neural network training problem (medium noise) for neural networks with $3$, $4$ and $5$ hidden nodes}\label{tab:BACKNM}
\begin{center}
\begin{tiny}
\begin{tabular}{c|l@{$\,\pm\,$}r|c|l@{$\,\pm\,$}r|c|l@{$\,\pm\,$}r|c}
\hline\hline
Optimizer & \multicolumn{3}{c}{3 hidden nodes} & \multicolumn{3}{|c}{4 hidden nodes} & \multicolumn{3}{|c}{5 hidden nodes}\\
\hline
MS-CAP  &$\mathbf{1.48e-01}$&$\mathbf{1.72e-06}$&&$\mathbf{1.44e-01}$&$\mathbf{8.16e-06}$&&$\mathbf{1.40e-01}$&$\mathbf{3.25e-06}$&\\
EBP    &$1.53e-01$&$4.30e-04$&+&$1.53e-01$&$4.30e-04$&+&$1.53e-01$&$5.53e-04$&+\\
RP    &$1.53e-01$&$3.36e-17$&+&$1.53e-01$&$2.78e-17$&+&$1.53e-01$&$3.25e-17$&+\\
\hline\hline
\end{tabular}
\end{tiny}
\end{center}
\end{table*}

\begin{table*}
\caption{Average MSE $\pm$ standard deviation and statistical comparison (reference: MS-CAP) for MS-CAP against EBP and RP on the neural network training problem (high noise) for neural networks with $3$, $4$ and $5$ hidden nodes}\label{tab:BACKNH}
\begin{center}
\begin{tiny}
\begin{tabular}{c|l@{$\,\pm\,$}r|c|l@{$\,\pm\,$}r|c|l@{$\,\pm\,$}r|c}
\hline\hline
Optimizer & \multicolumn{3}{c}{3 hidden nodes} & \multicolumn{3}{|c}{4 hidden nodes} & \multicolumn{3}{|c}{5 hidden nodes}\\
\hline
MS-CAP  &$1.50e-01$&$5.37e-15$&&$\mathbf{1.45e-01}$&$\mathbf{5.59e-15}$&&$\mathbf{1.44e-01}$&$\mathbf{7.70e-15}$&\\
EBP    &$\mathbf{1.46e-01}$&$\mathbf{1.54e-05}$&-&$1.46e-01$&$1.17e-05$&+&$1.46e-01$&$2.18e-05$&+\\
RP    &$1.46e-01$&$4.83e-17$&-&$1.46e-01$&$2.69e-17$&+&$1.46e-01$&$5.28e-17$&+\\
\hline\hline
\end{tabular}
\end{tiny}
\end{center}
\end{table*}

\section{Conclusions}\label{sec:conclusions}
\noindent
This paper proposes a Memetic Computing structure in which a population of candidate solutions, termed here \emph{coevolving aging particles}, are perturbed, independently, along each dimension. The perturbation occurs in such a way that the search radius progressively decays, based on the ``age'' of the particle, while the same particle is attracted with a progressively increasing force towards the current best solution in the swarm. When this logics fails at finding a fitness improvement, an evolutionary component is activated. The latter, employing a randomized pool of multiple mutation/recombination strategies typically used in Differential Evolution, attempts to further exploit the current genetic material and possibly reach unexplored areas of the search space. 

The proposed algorithm, named Multi-Strategy Coevolving Aging Particles (MS-CAP), has been tested over a diverse testbed in various dimensions ranging from $10$ to $1000$ and compared against ten modern meta-heuristics representing the-state-of-the art in continuous optimization. This comparison, assessed through a thorough statistical analysis, showed that the MS-CAP algorithm is superior on the employed setup to the state-of-the-art algorithms considered in this study, displaying a high performance in various landscapes characterized by different features in terms of multi-modality, separability, ill-conditioning, and dimensionality. 

To further demonstrate the efficacy and robustness of our approach, we presented an application of MS-CAP as a training algorithm for a Feedforward Neural Network in a robotics case study. Also in this case MS-CAP showed a very competitive performance in comparison with both state-of-the-art general-purpose meta-heuristics and classic training algorithms such as Error Back Propagation and Resilient Propagation. 

Future research will attempt to improve the proposed scheme, for example integrating it with local search logics or endowing it with self-adapting capabilities, and apply it to different real-world problems.

\nonumsection{Acknowledgments}
\noindent
INCAS\textsuperscript{3} is co-funded by the Province of Drenthe, the Municipality of Assen, the European Fund for Regional Development and the Ministry of Economic Affairs, Peaks in the Delta. This research is supported by the Academy of Finland, Akatemiatutkija 130600, Algorithmic Design Issues in Memetic Computing. The numerical experiments have been carried out on the computer network of the De Montfort University by means of the software for distributed optimization Kimeme\cite{web:kimeme} within the MemeNet Project. 

%

\nonumsection{References}
\bibliographystyle{ieeetr}
\bibliography{biblio_master}

%
%
%
%


%
\end{document}